\pdfoutput=1
\documentclass[11pt]{article}
\usepackage[final]{acl2023}
\providecommand{\nolinenumbers}{}
\providecommand{\linenumbers}{}
\usepackage{times}
\usepackage{latexsym}
\usepackage[T1]{fontenc}
\usepackage[utf8]{inputenc}
\usepackage{microtype}
\usepackage{inconsolata}
\usepackage{graphicx}
\usepackage{booktabs}
\usepackage{array}
\usepackage{multirow}
\usepackage{amsmath}
\usepackage{xcolor}
\usepackage{float}
\usepackage{placeins}
\usepackage{capt-of}
\usepackage{url}
\hypersetup{hidelinks}
\newcommand{\cellbg}[2]{\begingroup\setlength{\fboxsep}{1.5pt}\colorbox{#1}{\parbox[t]{\dimexpr\linewidth-3pt\relax}{\raggedright #2}}\endgroup}

\usepackage[normalem]{ulem}
\definecolor{suggadd}{HTML}{B00020}
\definecolor{suggdel}{HTML}{9A6A00}
\definecolor{suggmove}{HTML}{1A56B0}
\definecolor{suggacc}{HTML}{1B6E20}

\newcommand{\blfootnote}[1]{\begingroup\renewcommand\thefootnote{}\footnote{#1}\addtocounter{footnote}{-1}\endgroup}
\usepackage{transparent}
\definecolor{hlprev}{HTML}{B9E6B9}

\title{Ambiguous Examples Help, But Full Data Wins:\ Revisiting Dataset Cartography for Extractive Adversarial QA}

\author{Paul K. Mandal\textsuperscript{\dag,\ddag,$\diamondsuit$} \\
  \textsuperscript{\dag}Neurint LLC, Baton Rouge, LA \\
  \textsuperscript{\ddag}U.S. Army Cyber Corps, U.S. Army Reserve, USA \\
  \textsuperscript{$\diamondsuit$}Northwestern State University of Louisiana, Natchitoches, LA \\
  \texttt{paul@research.neurint.ai}
}

\begin{document}
\maketitle
\pagestyle{plain}
\thispagestyle{plain}
\raggedbottom

\begin{abstract}
Dataset cartography argues that training dynamics can reveal regions of a dataset that matter differently for learning: easy-to-learn examples aid optimization, hard-to-learn examples often expose annotation problems, and ambiguous examples contribute strongly to out-of-distribution generalization. A striking implication of the original work is that a model trained on only 33\% ambiguous examples can outperform full-data training on several OOD evaluations, including a QNLI reformulation of SQuAD evaluated on Adversarial SQuAD. We revisit this claim in a stricter setting: open-ended extractive question answering, where models must predict answer spans rather than classify question--sentence pairs. For each encoder family, we compute clean-source SQuAD training dynamics, train on full, random, easy-ranked, ambiguous-ranked, and hard-ranked subsets, and evaluate robustness on adversarial-only AddSent and AddOneSent rows. Ambiguous-ranked 33\% subsets beat random 33\% subsets on exact-match accuracy in all 14 model--evaluation cells and on F1 in 12/14 cells. However, full-data training beats ambiguous-ranked training in 12/14 exact-match cells and 12/14 F1 cells, with mean full-minus-ambiguous gains of 3.40 EM and 4.10 F1. The most thoroughly replicated ELECTRA-small panel is also the main exception to the ambiguous-over-random F1 gains. These results show that clean-source cartography remains useful for reduced-data selection, but its OOD advantage does not carry over as a general replacement for full-data training in extractive adversarial QA.
\end{abstract}

\section{Introduction}

Large supervised NLP datasets make it possible to train high-performing models, but high benchmark accuracy does not guarantee that a model has learned the intended task. Work on natural language inference has shown that models can exploit annotation artifacts and syntactic heuristics rather than robust semantic reasoning \citep{gururangan2018annotation,feng2019misleading,mccoy-etal-2019-right}. Reading comprehension raises the same issue in a different form. A system may locate plausible answer spans by relying on lexical overlap, local sentence matching, answer position, or question-insensitive patterns rather than by resolving the question against the passage \citep{jia-adversarial,kaushik-lipton-2018-much,mudrakarta-etal-2018-did,sugawara-etal-2018-makes}.

Dataset cartography offers a principled way to inspect such datasets. \citet{dataset_cartography} propose data maps: model-dependent visualizations that place each training example according to confidence in the gold label and variability of that confidence across epochs. The resulting maps reveal easy-to-learn, ambiguous, and hard-to-learn regions. The original paper's most important claim is not merely descriptive. In several classification-style settings, training on only 33\% ambiguous examples improves out-of-distribution (OOD) performance and can even outperform full-data training. Its QNLI experiment is especially relevant for us: QNLI is derived from SQuAD, but recasts question answering as sentence-level classification, and the ambiguous 33\% subset outperforms 100\% training on an Adversarial SQuAD-derived OOD evaluation \citep{dataset_cartography,wang2018glue}.

This paper asks whether that selection result survives the original SQuAD task formulation.\blfootnote{\raggedright\hspace*{-1.8em}Accepted to \emph{Findings of the Association for Computational Linguistics: EACL 2027}.}\blfootnote{\raggedright\rule{0pt}{13pt}\hspace*{-1.8em}Code available at \href{https://github.com/PaulKMandal/dataset-cartography-qa}{\texttt{https://github.com/}\newline\texttt{PaulKMandal/dataset-cartography-qa}}} In QNLI, the model decides whether a candidate sentence contains the answer. In extractive SQuAD, the model must choose the answer span from a passage. The difference matters because AddSent and AddOneSent attack span selection directly: they add plausible distractor sentences while leaving the correct answer unchanged \citep{jia-adversarial}. A training example that is ambiguous for sentence-level answer-containment classification need not teach a model to rank spans correctly when a distractor contains a tempting alternative answer. Conversely, full SQuAD training may expose the model to a broader range of answer types, answer positions, and lexical contexts that are useful for span-level robustness.

We therefore evaluate clean-source cartographic selection for extractive adversarial QA. For each model family, we compute a SQuAD data map using that model's extractive-QA training dynamics, select easy-ranked, ambiguous-ranked, and hard-ranked 33\% subsets (ranked by descending confidence, descending variability, and ascending confidence, respectively), and compare them with repeated random 33\% subsets and full-data training. We evaluate seven encoder families: ELECTRA-small, BERT-base, ELECTRA-base, RoBERTa-base, DistilBERT, ALBERT-base, and DeBERTa-v3-base. Our primary robustness metrics are adversarial-only exact match and token F1 on AddSent and AddOneSent. We also report paired failure rates, which ask whether a model that answers the original row correctly fails after adversarial perturbation.

The empirical picture is two-sided. Ambiguous-ranked examples are genuinely useful reduced-data selections: they beat random 33\% on adversarial-only exact-match accuracy in all 14 model--evaluation cells and on F1 in 12 of 14 cells. But the stronger ``ambiguous beats full'' pattern from the QNLI-style cartography result does not generally hold. Full-data training is best in 12 of 14 exact-match cells and 12 of 14 F1 cells. ELECTRA-small, the most thoroughly replicated panel, is also the main case where ambiguous-ranked F1 underperforms random. Thus the main result is a boundary on the selection claim: ambiguous examples help reduced-data robustness, but full clean-source training usually wins for open-ended extractive adversarial QA.

Our contributions are threefold. First, we provide a multi-model evaluation of clean-source cartographic selection for extractive adversarial QA. Second, we show that the strongest OOD selection claim does not reproduce under the original extractive formulation: the full-data replacement effect reported for QNLI-style classification does not carry over to extractive span prediction in our experiments. Third, we supply data maps, subset-overlap diagnostics, repeated random baselines, model-specific result profiles, paired robustness analyses, and reproducibility artifacts that clarify what clean SQuAD training dynamics do and do not buy.

\section{Background and Related Work}

\paragraph{Dataset cartography and training dynamics.}
Dataset cartography builds on the idea that examples can be characterized by how a model behaves on them during training. \citet{dataset_cartography} define confidence as the mean probability assigned to the true label across epochs, variability as the standard deviation of that probability, and correctness as the fraction of epochs with a correct prediction. Their data maps reveal three broad regions: easy-to-learn examples with high confidence and low variability, ambiguous examples with high variability, and hard-to-learn examples with low confidence. The original paper reports three qualitative roles for these regions: ambiguous examples are useful for OOD generalization, easy examples help optimization, and hard examples often surface label errors.

A broader literature also scores or selects examples using learning behavior. Forgetting-event methods track examples that transition from correct to incorrect during training \citep{toneva2018empirical}. Active-bias methods emphasize high-variance examples during optimization \citep{chang2017active}. Curriculum and self-paced learning order training by difficulty \citep{bengio2009curriculum,kumar2010self}. Influence functions and margin-based criteria can identify harmful or mislabeled examples \citep{koh2017understanding,pleiss2020identifying}. AFLite adversarially filters predictable examples to reduce measurable dataset biases \citep{lebras2020adversarial}. Our work follows the hard-selection diagnostic setup of dataset cartography: we train models from scratch on ranked subsets to test what each ranked selection contributes. Dataset cartography has also seeded a sizable downstream literature. Data maps directly inform active learning \citep{zhang-plank-2021-cartography-active}; training-dynamics scores drive dynamic data pruning for NLU fine-tuning \citep{attendu-corbeil-2023-nlu} and data pruning for LLM pretraining at scale \citep{marion2023less}; training trajectories of small proxy models select fine-tuning data for larger models \citep{yang2024smalltolarge}; and recent surveys organize this data-selection literature \citep{albalak2024survey}. Much of this work inherits the assumption our study bounds: that subsets chosen from clean-source training dynamics retain the strengths of full-data training.

\paragraph{Question answering artifacts and shortcut behavior.}
Extractive QA benchmarks are designed to test whether systems can locate an answer span from a passage \citep{squad}. However, several studies show that high performance can coexist with shortcut behavior. \citet{kaushik-lipton-2018-much} evaluate how much reading is required by popular RC benchmarks and find that partial-input baselines can perform surprisingly well on some datasets, even though SQuAD is better constructed than several alternatives. \citet{mudrakarta-etal-2018-did} use attribution to show that QA models can ignore important question terms and use this behavior to craft adversarial perturbations. \citet{sugawara-etal-2018-makes} analyze what makes MRC questions easy and show that superficial cues can inflate apparent progress. These findings motivate tests that preserve the answer but perturb the evidence context.

\paragraph{Adversarial and challenge-set QA evaluation.}
AddSent and AddOneSent were introduced to test reading-comprehension systems by inserting distractor sentences into SQuAD passages without changing the correct answer \citep{jia-adversarial}. This evaluation probes whether a model can prioritize the original answer evidence over a plausible distractor. Later QA datasets and evaluations broaden the same concern. SQuAD 2.0 adds adversarially written unanswerable questions that require abstention \citep{rajpurkar2018know}; DROP requires discrete reasoning over paragraphs \citep{dua2019drop}; MRQA evaluates cross-dataset generalization across unified RC datasets \citep{fisch2019mrqa}; contrast sets perturb examples to expose local decision-boundary brittleness \citep{gardner-etal-2020-evaluating}; and adversarial human annotation creates questions specifically designed to fool a model in the loop \citep{bartolo2020beat}. Our study is narrower than this full landscape: we use AddSent/AddOneSent as controlled tests of span-selection robustness under distractor perturbation.

\paragraph{QNLI, extractive QA, and task formulation.}
QNLI is derived from SQuAD but converts the problem into sentence-pair classification: the model predicts whether a context sentence contains the answer to a question \citep{wang2018glue}. This formulation is valuable for broad language-understanding benchmarks, but it removes the final span-selection problem. In extractive SQuAD, a model must score start and end positions and produce an answer string. The distinction is central to our paper. Dataset cartography's QNLI result shows that ambiguous sentence-pair examples can improve Adversarial SQuAD-style OOD accuracy. We test whether the same kind of clean-source selection provides a substitute for full training when the model must make span predictions under AddSent/AddOneSent distractors.

\paragraph{Clean-source selection vs. challenge-domain inoculation.}
Our setting is also distinct from adversarial fine-tuning and challenge-domain inoculation. In inoculation-style experiments, a model is exposed to a small number of examples from the challenge distribution to improve performance on related challenge examples. Cartographic inoculation applies the cartography idea to choose adversarial or challenge examples for fine-tuning \citep{chen2024cartographic}. That asks how to select data when the target challenge distribution is already available. Our question is stricter: whether dynamics computed only on clean SQuAD training examples identify source examples whose use improves AddSent/AddOneSent robustness.

\begin{figure*}[t]
\centering
\includegraphics[width=.92\linewidth]{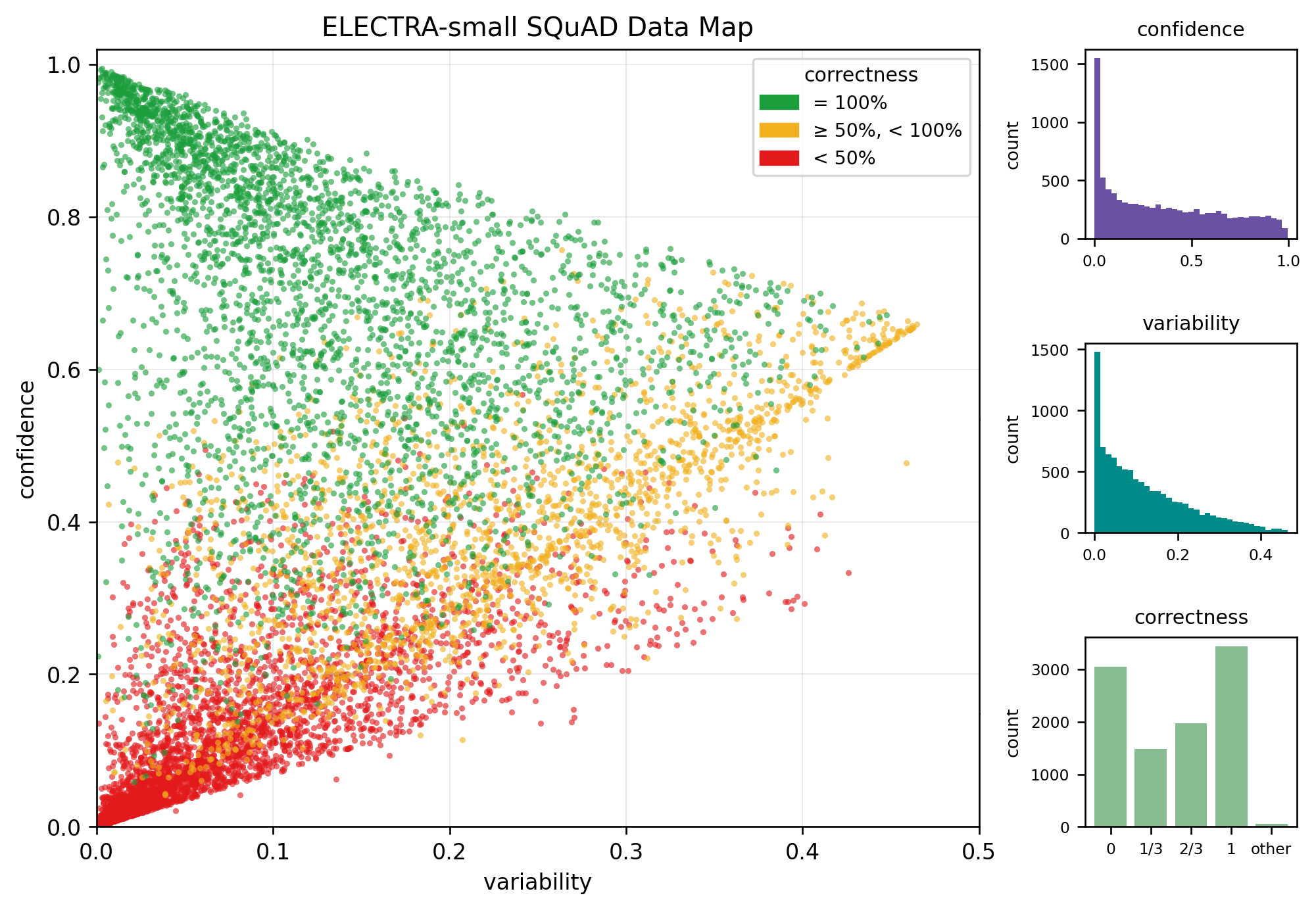}
\caption{Data map for clean SQuAD training examples under the ELECTRA-small extractive-QA model. The x-axis shows variability of gold-span confidence and the y-axis shows mean gold-span confidence. For clarity, the scatter plot and side distributions use a deterministic 10K sample of map points. Right-side plots show confidence, variability, and correctness distributions. Points are colored by exact-match correctness: green marks examples always answered correctly, amber at least half but not all, and red less than half. In the terms of \citet{dataset_cartography}, high-confidence, low-variability examples (upper left) are easy-to-learn, high-variability examples (right) are ambiguous, and low-confidence examples (lower left) are hard-to-learn. The side plots retain the continuous confidence and variability distributions and the discrete correctness distribution; correctness is discrete because most grouped SQuAD examples have one recorded prediction per epoch in a three-epoch dynamics run.}
\label{fig:squad-datamap}
\end{figure*}

\paragraph{Relation to our prior work.}
Our prior work \citep{mandal2025cartography} studied dataset cartography for extractive SQuAD and reported that training-dynamics subsets do not improve ELECTRA-small robustness against Adversarial SQuAD relative to random controls. That result is a useful warning that cartographic selection may not transfer cleanly from QNLI-style classification to span prediction, but it leaves open whether the effect is architecture-dependent, whether adversarial-only filtering changes the conclusion, and whether ambiguous subsets can replace full-data training across model families. The present study addresses those gaps by expanding to seven encoder families, computing model-specific source maps, separating adversarial-only AddSent/AddOneSent rows, comparing against repeated random baselines and full-data training, and reporting EM/F1, paired robustness, and reproducibility analyses.

\paragraph{Experimental reliability.}
Small differences between data-selection methods can be unstable under random initialization, data order, and subset sampling. \citet{Dodge:2020} show that fine-tuning pretrained language models can vary substantially with initialization and training order, and \citet{dodge2019show} argue for better reporting of experimental results. We therefore treat random 33\% training as a distribution over persistent subset draws and training seeds rather than as a single baseline point.

\section{Method}

\subsection{Training dynamics for extractive QA}

The original cartography definition assumes a classification model with a probability assigned to the gold class. Extractive QA instead predicts start and end positions. We instantiate confidence as the joint probability assigned to the gold span, averaged across training-dynamics records. Variability is the standard deviation of this gold-span confidence, and correctness is the fraction of records in which the predicted answer exactly matches a gold answer. The raw logs store the within-epoch training position for each recorded example, but most grouped SQuAD examples are observed once per epoch in a three-epoch run; consequently, correctness is coarser than confidence and variability, with most mass at 0, 1/3, 2/3, and 1. These coordinates define a SQuAD extractive-QA data map with the same conceptual axes as the original framework.

Figure~\ref{fig:squad-datamap} shows the ELECTRA-small SQuAD data map as the primary illustrative map, and Appendix~\ref{app:model-profiles} provides the corresponding maps for all seven encoders. The maps follow the broad cartographic geometry of the original work: high-confidence stable examples occupy an easy region, high-variability examples define an ambiguous band, and low-confidence examples form a hard region. Unlike QNLI, these maps are computed for span prediction rather than sentence-level answer-containment classification. Appendix~\ref{app:aggregate} reports the model-specific map statistics.

\subsection{Ranked source subsets}

Following the selection logic of \citet{dataset_cartography}, we use ranked top-$k$ selections rather than disjoint geometric partitions. The easy-ranked subset selects high-confidence examples, the ambiguous-ranked subset selects high-variability examples, and the hard-ranked subset selects low-confidence examples. Each subset contains 33\% of the SQuAD training set, selected by a global ranking over 87,599 grouped examples. Because these are ranked selections, an example can belong to more than one selected subset; for example, a high-confidence example can also have high variability. The visualization draws one fixed 10,000-example sample per encoder (random state 1009) and highlights sampled IDs belonging to each global subset; the highlighted counts need not be equal. Appendix~\ref{app:aggregate} audits this overlap.

This distinction is important for interpreting the subset overlays in Appendix~\ref{app:model-profiles}. We distinguish the correctness-category colors, qualitative neighborhoods in the confidence--variability plane, and globally ranked training subsets. The ranked subsets are the experimental intervention; neither a color nor a neighborhood alone establishes training-subset membership.

\subsection{Training and evaluation}
\label{sec:training-eval}

We compare five training conditions: full SQuAD training, random 33\%, easy-ranked 33\%, ambiguous-ranked 33\%, and hard-ranked 33\%. The 33\% budget matches the original cartography selection experiments and controls for the amount of training data in reduced-data comparisons. ELECTRA-small and BERT-base use 10 persistent random draws crossed with three training seeds. The remaining model panels use five persistent random draws crossed with three training seeds. ELECTRA-small and BERT-base were the original canonical panels and were replicated more extensively; the remaining panels use five draws for compute reasons. The asymmetry affects only the precision of the random-33\% baseline estimate; the within-model full-vs-ambiguous comparisons do not depend on the number of draws. Deterministic selected subsets are trained with three seeds. All models are trained for three epochs.

We evaluate seven encoder families: ELECTRA-small, BERT-base, ELECTRA-base, RoBERTa-base, DistilBERT, ALBERT-base, and DeBERTa-v3-base \citep{electra,devlin-etal-2019-bert,liu2019roberta,sanh2019distilbert,lan2020albert,he2023debertav3}. The multi-model panel tests whether a clean-source cartographic selection signal transfers across model families when the task remains extractive QA.

The released AddSent/AddOneSent-style files contain both original rows and adversarial rows. Reporting all rows would blend clean and adversarial examples, so our primary robustness metrics use adversarial-only rows. Appendix~\ref{app:aggregate} reports the row composition. We also compute paired robustness on the 787 base questions for which both original and adversarial variants are available. A paired failure occurs when a model answers the original row correctly but fails on at least one adversarial variant.

\paragraph{Reproducibility.}
All experiment commands, per-run YAML configurations, training metrics, evaluation metrics, subset assignments, environment snapshots, and raw source-run training dynamics were retained for audit. Code, metrics, and reproduction artifacts are available at \url{https://github.com/PaulKMandal/dataset-cartography-qa}. The logged-artifacts inventory in Appendix~\ref{app:repro} states where the scripts, per-run configurations, subset definitions, and evaluation summaries reside within that repository. Appendix~\ref{app:repro} reports the execution environment, hyperparameters, run counts, and computational budget; Appendix~\ref{app:responsible} addresses responsible-NLP checklist items relevant to this study.

\section{Results}

\begin{table*}[t]
\noindent\begin{minipage}{\textwidth}

\centering
\small
\setlength{\tabcolsep}{3pt}
\renewcommand{\arraystretch}{1.15}
\begin{tabular}{p{.28\linewidth}p{.23\linewidth}rrrr}
\toprule
Study / task formulation & Evaluation & Full & Random 33\% & Ambiguous 33\% & Amb.-Full \\
\midrule
Swayamdipta et al. QNLI (sentence classification) & Adv. SQuAD-derived acc. & 81.7 & 78.3 & \textbf{83.9} & +2.2 \\
This work, extractive QA & AddSent EM & \textbf{55.64} & 49.05 & 51.90 & -3.74 \\
This work, extractive QA & AddOneSent EM & \textbf{55.22} & 48.85 & 51.93 & -3.29 \\
This work, extractive QA & AddSent F1 & \textbf{61.73} & 55.35 & 56.87 & -4.86 \\
This work, extractive QA & AddOneSent F1 & \textbf{61.74} & 55.16 & 56.94 & -4.80 \\
\bottomrule
\end{tabular}
\captionof{table}{The central contrast with the original data-selection result. Swayamdipta et al. report that 33\% ambiguous training beats full-data training on QNLI evaluated with an Adversarial SQuAD-derived OOD set. In extractive QA, ambiguous-ranked subsets improve over random reduced-data training but do not generally beat full-data training. Our rows report model-averaged adversarial-only means across the seven canonical panels.}
\label{tab:sway-contrast}

\end{minipage}

\end{table*}
\begin{figure*}[t]
\input{tables/fixed_main_fig2.tex}
\end{figure*}
\begin{table*}[t]
\noindent\begin{minipage}{\textwidth}

\centering\small\setlength{\tabcolsep}{4.2pt}
\begin{tabular}{lrrrrrr}
\toprule
 & \multicolumn{3}{c}{AddSent EM} & \multicolumn{3}{c}{AddOneSent EM} \\
Model & Full & Random & Ambiguous & Full & Random & Ambiguous \\
\midrule
ELECTRA-small & \textbf{42.60 $\pm$ 1.23} & 34.08 $\pm$ 1.06 & 34.22 $\pm$ 0.85 & \textbf{42.57 $\pm$ 1.16} & 33.81 $\pm$ 1.16 & 34.43 $\pm$ 1.13 \\
BERT-base & \textbf{51.00 $\pm$ 0.23} & 44.30 $\pm$ 1.28 & 47.42 $\pm$ 1.35 & \textbf{50.49 $\pm$ 0.60} & 43.91 $\pm$ 1.34 & 47.73 $\pm$ 2.04 \\
ELECTRA-base & \textbf{62.57 $\pm$ 0.78} & 54.85 $\pm$ 1.60 & 59.45 $\pm$ 1.39 & \textbf{62.35 $\pm$ 0.32} & 54.36 $\pm$ 1.92 & 59.81 $\pm$ 1.20 \\
RoBERTa-base & \textbf{61.25 $\pm$ 0.52} & 57.76 $\pm$ 1.45 & 59.90 $\pm$ 0.79 & \textbf{62.13 $\pm$ 0.13} & 57.81 $\pm$ 1.31 & 59.76 $\pm$ 0.57 \\
DistilBERT & \textbf{46.08 $\pm$ 1.14} & 38.34 $\pm$ 1.41 & 40.08 $\pm$ 1.22 & \textbf{45.70 $\pm$ 1.48} & 37.78 $\pm$ 1.27 & 40.24 $\pm$ 1.27 \\
ALBERT-base & \textbf{55.40 $\pm$ 1.04} & 49.77 $\pm$ 1.50 & 52.90 $\pm$ 2.06 & \textbf{55.65 $\pm$ 0.92} & 49.99 $\pm$ 1.45 & 53.66 $\pm$ 1.73 \\
DeBERTa-v3-base & 74.67 $\pm$ 0.90 & 71.84 $\pm$ 0.94 & \textbf{74.99 $\pm$ 1.11} & 75.01 $\pm$ 0.73 & 71.83 $\pm$ 1.05 & \textbf{75.35 $\pm$ 1.44} \\
\bottomrule
\end{tabular}
\captionof{table}{Adversarial-only exact-match results for the three central training conditions. Values are means $\pm$ standard deviations over training seeds and random subset draws where applicable. Bold marks the best of the three shown conditions for each model and evaluation. Full tables including easy-ranked and hard-ranked conditions, plus token-F1 results, appear in Appendix~\ref{app:aggregate}.}
\label{tab:main-em}

\end{minipage}

\end{table*}

\subsection{The central contrast: QNLI cartography vs. extractive QA}

The original dataset cartography paper reports that ambiguous examples are especially valuable for OOD generalization. In its QNLI experiment, 33\% ambiguous training reaches 83.9 accuracy on an Adversarial SQuAD-derived OOD test, compared with 81.7 for full-data training and 78.3 for random 33\% training. This is the strongest comparison point for our work because QNLI is derived from SQuAD but recasts QA as sentence-level classification.

Our extractive-QA results draw a different boundary. Ambiguous 33\% selection is still useful: it beats random 33\% on exact-match accuracy in every adversarial-only model/evaluation cell and on F1 in 12 of 14 cells. But it does not generally beat full data. Full-data training beats ambiguous 33\% in 12 of 14 exact-match cells and 12 of 14 F1 cells, with mean full-minus-ambiguous gains of 3.40 EM and 4.10 F1. Table~\ref{tab:sway-contrast} summarizes this contrast. Figure~\ref{fig:effect-summary} visualizes the same boundary in exact-match terms, separating the reduced-data gain over random from the full-data advantage over ambiguous selection. Detailed single-evaluation versions are in Appendix~\ref{app:aggregate}.

\subsection{Exact-match accuracy results}

Exact-match accuracy is central for extractive QA because it asks whether the predicted answer span exactly matches a gold answer. Table~\ref{tab:main-em} summarizes the three central conditions. Ambiguous-ranked training improves over random 33\% in every model/evaluation cell. The mean ambiguous-minus-random exact-match gain is 2.82 points. This is a genuine positive result for clean-source cartography as reduced-data selection. Full five-condition tables are in Appendix Tables~\ref{tab:addsent-em} and~\ref{tab:addonesent-em}.

However, the stronger ``ambiguous beats full'' pattern does not hold. Full-data training is best in 12 of 14 exact-match cells. The only exceptions are DeBERTa-v3-base on AddSent and AddOneSent, where ambiguous-ranked 33\% slightly exceeds full-data training. These differences are within run-to-run variability, and random 33\% training already approaches full-data performance for this model, so we read the comparison as a near-tie.

\subsection{Token-F1 results}

F1 tells the same story with a slightly weaker ambiguous-over-random pattern. Appendix Tables~\ref{tab:addsent-f1} and~\ref{tab:addonesent-f1} show that ambiguous-ranked 33\% beats random 33\% in 12 of 14 F1 cells, with a mean gain of 1.85 points. Full-data training is again best in 12 of 14 cells and exceeds ambiguous-ranked training by a mean of 4.10 F1 points.

The two ambiguous-over-random F1 failures are both in ELECTRA-small: AddSent and AddOneSent. ELECTRA-small is the model for which our random baseline is most extensive, with 10 random draws crossed with three training seeds. The deficits are within run-to-run variability, so we read them simply as the absence of a reliable F1 gain: ambiguous clean-source examples are often useful, but the effect is not universal. It also cautions against drawing broad cartography conclusions from a single small encoder. The clean-domain ordering matches the adversarial one: Appendix Table~\ref{tab:clean-squad} reports original SQuAD-dev EM and F1 for the three central conditions.

\subsection{Easy and hard subsets are not robust alternatives}

The original cartography paper reports that hard-to-learn examples can also improve OOD performance in some classification settings and that hard examples often expose labeling errors. Our extractive-QA results do not support hard-ranked training as a general robustness strategy. Top-confidence (easy-ranked) subsets never beat random 33\% on exact-match accuracy or F1. Low-confidence (hard-ranked) subsets never beat random 33\% on exact-match accuracy and beat random on F1 in only 2 of 14 cells. In extractive QA, hard examples may reflect span-localization difficulty, annotation edge cases, long contexts, or model limitations rather than adversarially useful training signal. We therefore do not equate hard SQuAD examples with label errors without a separate human audit.

\begin{figure*}[t]
\centering
\includegraphics[width=.96\linewidth]{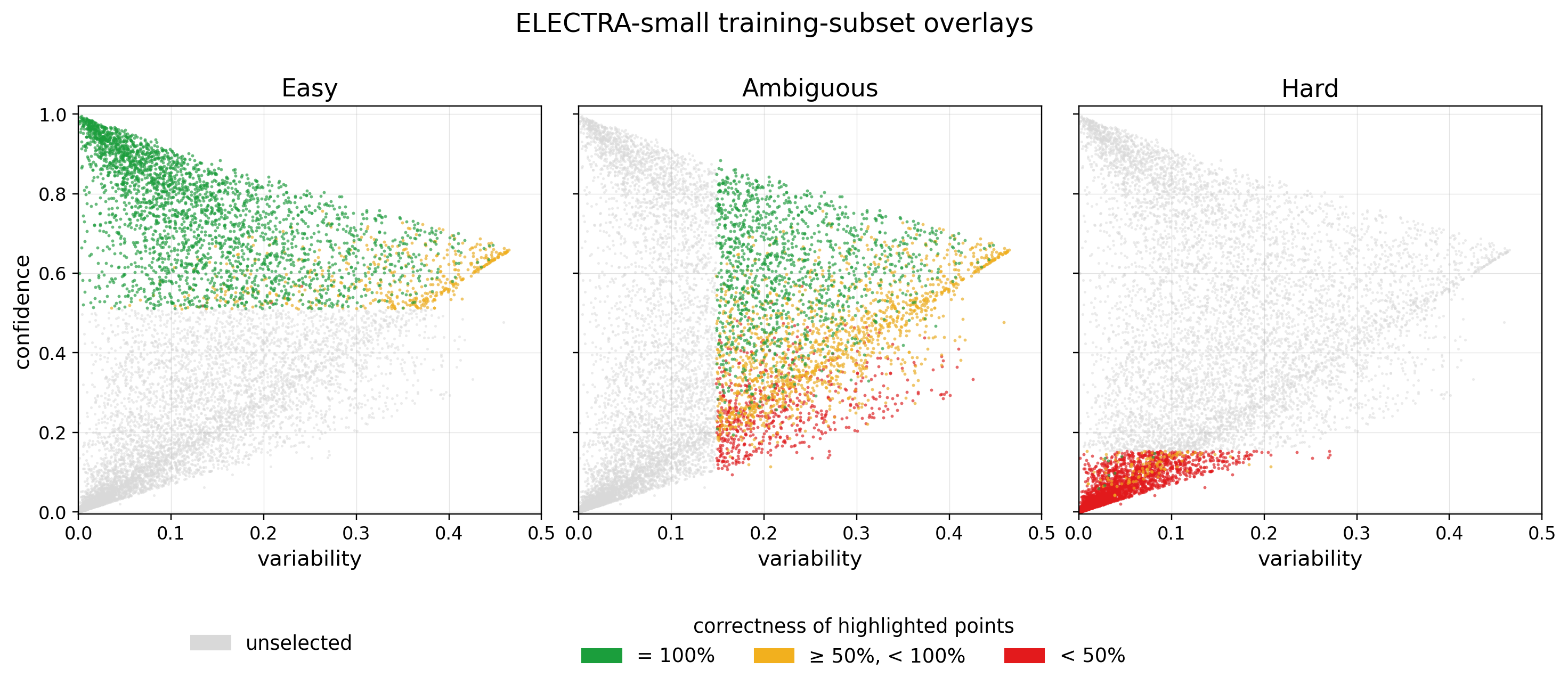}
\caption{Easy, Ambiguous, and Hard 33\% training-subset overlays for ELECTRA-small on the model's SQuAD data map. The same deterministic 10K map sample is used in all three panels. Gray points show the fixed sample; highlighted points are those sampled examples that belong to the corresponding globally ranked 33\% training subsets. Highlight colors give each example's correctness category, so multiple colors can occur in any panel. The three rankings are computed independently, so the panels are not disjoint: an example selected by more than one ranking is highlighted in each corresponding panel; Table~\ref{tab:subset-overlap} quantifies the overlaps.}
\label{fig:selected-subsets}
\end{figure*}

\subsection{Paired robustness}

The paired analysis asks a stricter question: among questions a model can answer before perturbation, how often does it fail after the distractor is added? Appendix Table~\ref{tab:paired-failure} shows that ambiguous-ranked selection has a lower paired failure rate than random 33\% in 13 of 14 cells, with a mean failure-rate reduction of 2.44 percentage points. This supports the exact-match and F1 results: ambiguous clean-source examples provide a reduced-data robustness signal, even though full-data training usually remains stronger.

\section{Data-Map Diagnostics}

The selection results are easier to interpret when viewed against the data maps. Figure~\ref{fig:selected-subsets} shows the ELECTRA-small global-subset overlays: a single deterministic 10K map sample is used for all panels, and sampled example IDs are intersected with each globally ranked 33\% training subset. The panels therefore retain the actual globally selected subset membership rather than re-ranking the sample to force equal highlight counts. Appendix~\ref{app:aggregate} provides aggregate audits, full result tables, master visualizations, and global-subset intersection audits; Appendix~\ref{app:qualitative} contains qualitative examples; Appendix~\ref{app:model-profiles} groups the model-specific subset overlays by encoder family. The key methodological point is that easy-, ambiguous-, and hard-ranked are global top-$k$ selection rules, whereas green/amber/red refer to correctness categories; neither corresponds to a mutually exclusive geometric partition of the map.

\section{Discussion}

\paragraph{The main result is a boundary on the original selection claim.}
The original dataset cartography result is powerful because ambiguous subsets can outperform full-data training in several OOD classification-style settings. Our results show that this does not generally transfer from QNLI-style sentence classification to extractive SQuAD-style span prediction. In true extractive QA, where the model must choose an answer span under distractor perturbation, full-data training usually wins. This is not a negative result about cartography as a diagnostic tool or reduced-data selector. It is a boundary on the claim that ambiguous subsets can replace full-data training for OOD robustness. Because capacity, architecture, pretraining objective, absolute task skill, and optimization covary across the seven encoders, the panel does not isolate task formulation as the mechanism behind this boundary.

\paragraph{Ambiguous-ranked examples remain a useful reduced-data selection.}
Ambiguous-ranked subsets beat repeated random baselines on exact-match accuracy in every cell and on F1 in most cells. This is a real signal. The right conclusion is therefore two-sided: ambiguous examples are better than random reduced-data selection, but not better than using all clean SQuAD training data for extractive adversarial QA. Because each data map is computed from the model's own full-data training dynamics, the selection signal is a diagnostic: identifying the subset already requires a full-data training pass, so it saves no training compute. Whether maps from a cheaper proxy model can select data for a larger target is a separate question we leave to future work.

\paragraph{ELECTRA-small shows no reliable ambiguous-F1 gain.}
ELECTRA-small is the one model for which ambiguous-ranked F1 underperforms random on both adversarial evaluations. The deficits are within run-to-run variability, so we do not interpret them further; with the panel's extensive random-baseline protocol, the absence of a gain is well estimated. Stronger or different encoder families often benefit from ambiguous selection, but ELECTRA-small does not show the same F1 gain.

\paragraph{Hard examples do not behave like adversarial training data.}
Hard-ranked subsets usually underperform random selection. This matters because hard examples in data maps can have multiple interpretations: mislabeled examples, annotation ambiguity, examples outside the model's capacity, or span-localization failures. For SQuAD, hard examples are not automatically adversarially informative. A human audit could test whether hard-ranked examples contain annotation problems, but the downstream robustness results alone do not support training on hard examples as a robust data-selection strategy.

\paragraph{Why task formulation may matter.}
QNLI asks whether a sentence contains an answer; extractive SQuAD asks which span in a context is the answer. AddSent/AddOneSent attacks exploit exactly this span-selection vulnerability by adding plausible distractors. A clean training example that is ambiguous for sentence-level classification need not teach the span-ranking behavior needed to ignore distractor answers. Conversely, full-data training may expose the model to a wider range of answer types, contexts, and span positions that matter for extractive robustness. This offers a plausible account of why our strongest result is not ``cartography fails,'' but ``cartography's full-data replacement effect did not reproduce under extractive span prediction.''
\section{Conclusion}

We revisited dataset cartography for adversarial question answering under a stricter extractive formulation. Compared with the original QNLI-style result, our multi-model span-prediction experiments show a different pattern. Ambiguous-ranked 33\% subsets are consistently better than random 33\% subsets for reduced-data robustness, especially in exact-match accuracy, but full-data training remains best in most AddSent/AddOneSent adversarial-only evaluations. The finding is therefore a clear boundary on the strongest interpretation of dataset cartography: ambiguous examples are valuable, but for extractive adversarial QA they do not generally replace the breadth of the full clean training set. These claims are limited to classical extractive QA on SQuAD v1.1 under AddSent/AddOneSent distractor evaluation.

\section{Limitations}

Our study focuses on SQuAD v1.1 and the AddSent/AddOneSent adversarial distractor setting. These benchmarks are useful controlled tests of span-selection robustness, but they do not cover unanswerable QA, multi-hop QA, long-context QA, or generative QA. We also study clean-source subset selection rather than adversarial fine-tuning; the conclusions do not rule out challenge-domain cartographic inoculation. Finally, although the data maps identify low-confidence hard examples, we do not perform a human label audit of hard SQuAD examples, so we do not claim that the hard-ranked subset primarily consists of annotation errors. All models also share one fixed fine-tuning protocol (learning rate, warmup, and epoch count), which keeps the within-model comparisons controlled but leaves cross-model differences entangled with optimization choices. Per-family learning-rate sweeps would be needed before reading differences in gap size across encoders as capacity or architecture effects.

\FloatBarrier
\bibliographystyle{acl_natbib}
\bibliography{custom}

\appendix
\clearpage
\section{Reproducibility and Computational Budget}
\label{app:repro}
\subsection{Execution environment}
We treated reproducibility as part of the experimental design. The experiment repository defines the software environment using a checked-in Nix flake and a uv-managed Python environment. The working branch contains \texttt{flake.nix}, \texttt{flake.lock}, \texttt{pyproject.toml}, and \texttt{uv.lock}. Our code is forked from \texttt{fp-dataset-artifacts} \citep{bostrom2021fpdatasetartifacts}. It has been substantially modified and expanded upon for this study. Each canonical panel records an environment snapshot, command log, per-run YAML configurations, training metrics, evaluation metrics, and source-run training dynamics. The logged canonical environments used Python 3.11.15, uv 0.9.30, Nix 2.34.7, and PyTorch 2.3.1+cu121.

The intended reproduction entry point is the GPU-server shell, \texttt{nix develop .\#server}, followed by \texttt{uv sync --frozen --extra cuda --group dev}. Local checks use the corresponding CPU environment. The environment table below records the versions that matter most for reproducing the reported training and evaluation runs.

\subsection{Run configuration and hyperparameters}
Each training run is represented by a YAML configuration file and a corresponding command-log entry. The configurations record the model identifier, training subset, subset fraction, random draw identifier when applicable, seed, data paths, output directory, confidence definition, number of epochs, maximum sequence length, learning rate, warmup ratio, weight decay, batch sizes, mixed-precision settings, and whether training dynamics were saved.

Across models, we used the same core QA fine-tuning settings. We did not tune hyperparameters on AddSent or AddOneSent. The goal of the experiment is a controlled comparison among full, random, easy-ranked, ambiguous-ranked, and hard-ranked source-training conditions rather than a search for the best model-specific hyperparameters.

\subsection{Logged artifacts}
The reproducibility package contains canonical panel directories, run configurations, command logs, environment snapshots, training metrics, evaluation metrics, final model weights, cartography scores, subset assignments, selected subset files, raw source-run training dynamics, figure-generation scripts, table-generation scripts, and package-verification scripts. For each canonical model family, the retained raw training dynamics can be re-aggregated by SQuAD example id to reproduce the corresponding cartography scores. In the released audit, the materialized cartography confidence column matches the mean of raw joint gold-span confidence.

\subsection{Computational budget}
All experiments were single-GPU fine-tuning and evaluation runs. ELECTRA-small, BERT-base, ELECTRA-base, RoBERTa-base, DistilBERT, and ALBERT-base were run on one NVIDIA RTX 4090. DeBERTa-v3-base was run on one NVIDIA A100. The canonical experiment set contains 219 training commands and 657 evaluation commands. Summing command-log elapsed times gives approximately 33.1 single-GPU hours for training and 13.2 single-GPU hours for evaluation, with a total of 46.3 logged command hours after including dynamics aggregation, subset selection, and small audit/materialization commands. The RunPod cost for the experiment suite was approximately \$70.

\clearpage
\nolinenumbers
\onecolumn
\noindent\begin{minipage}{\textwidth}

\centering\small\renewcommand{\arraystretch}{1.2}\setlength{\tabcolsep}{5pt}
\begin{tabular}{p{.16\linewidth}p{.32\linewidth}p{.42\linewidth}}
\toprule
Component & Logged value & Notes \\
\midrule
Python & 3.11.15 & Runtime recorded in canonical environment logs. \\
Nix & 2.34.7 & Used to enter the system shell for local and GPU-server execution. \\
uv & 0.9.30 & Used to sync the Python environment from the checked-in project definition. \\
PyTorch & 2.3.1+cu121 & CUDA 12.1 wheel used for GPU runs. \\
Transformers & 4.41.2 & Hugging Face implementation used for encoder QA fine-tuning. \\
Datasets / Evaluate / Accelerate & 2.19.2 / 0.4.2 / 0.30.1 & Used for data handling, metrics support, and Trainer execution. \\
Other pinned packages & NumPy 1.26.4;\newline pandas 2.2.2;\newline SciPy 1.13.1;\newline PyArrow 16.1.0;\newline PyYAML 6.0.1;\newline matplotlib 3.8.4;\newline NLTK 3.8.1 & Versions are pinned in the uv project definition. \\
\bottomrule
\end{tabular}
\captionof{table}{Software environment recorded for the canonical experiment package. The repository also contains checked-in \texttt{flake.nix}, \texttt{flake.lock}, \texttt{pyproject.toml}, and \texttt{uv.lock} files for reproducing the system and Python environments.}
\label{tab:environment}

\end{minipage}
\vspace{12pt}
\noindent\begin{minipage}{\textwidth}

\centering\small\renewcommand{\arraystretch}{1.2}\setlength{\tabcolsep}{4.5pt}
\begin{tabular}{p{.22\linewidth}p{.25\linewidth}p{.45\linewidth}}
\toprule
Setting & Value & Notes \\
\midrule
Task & Extractive QA & Models predict start and end positions over SQuAD-style contexts. \\
Training epochs & 3 & Used for full, random, easy-ranked, ambiguous-ranked, and hard-ranked conditions. \\
Maximum sequence length & 384 & QA tokenization length for training and evaluation. \\
Learning rate & $5\times10^{-5}$ & Fixed across panels; no evaluation-set tuning. \\
Warmup ratio & 0.0 & Fixed across panels. \\
Weight decay & 0.0 & Fixed across panels. \\
Per-device train batch size & 32 & Single-GPU training. \\
Per-device eval batch size & 64 & Single-GPU evaluation. \\
Precision & fp16 train/eval & Recorded in run configurations. \\
Training seeds & 13, 21, 42 & Used for deterministic selected subsets and full-data training. \\
Random subset draws & 10 or 5 & ELECTRA-small and BERT-base use 10 persistent draws; the remaining model panels use 5. \\
Cartography source run & Full-data seed 42 & Each model family builds its data map from its own source full-data training dynamics. \\
Primary confidence field & Joint gold-span probability & $p(\text{gold start})p(\text{gold end})$ aggregated by SQuAD example id. \\
Subset budget & 33\% of SQuAD train & Ranked easy, ambiguous, hard, and random subsets use the same data budget. \\
\bottomrule
\end{tabular}
\captionof{table}{Core hyperparameters and run settings. Each canonical run also has a per-run YAML configuration that records the model checkpoint, subset source, seed, data paths, output path, and training/evaluation arguments.}
\label{tab:hyperparameters}

\end{minipage}
\vspace{12pt}
\noindent\begin{minipage}{\textwidth}

\centering\small\setlength{\tabcolsep}{7pt}
\begin{tabular}{lrrrrr}
\toprule
Model & Random draws & \shortstack{Full/selected\\train runs} & \shortstack{Random\\train runs} & \shortstack{Total\\train runs} & Eval cmds \\
\midrule
ELECTRA-small & 10 & 12 & 30 & 42 & 126 \\
BERT-base & 10 & 12 & 30 & 42 & 126 \\
ELECTRA-base & 5 & 12 & 15 & 27 & 81 \\
RoBERTa-base & 5 & 12 & 15 & 27 & 81 \\
DistilBERT & 5 & 12 & 15 & 27 & 81 \\
ALBERT-base & 5 & 12 & 15 & 27 & 81 \\
DeBERTa-v3-base & 5 & 12 & 15 & 27 & 81 \\
\midrule
Total & -- & 84 & 135 & 219 & 657 \\
\bottomrule
\end{tabular}
\captionof{table}{Canonical run matrix. Full/selected train runs correspond to four deterministic conditions---full, easy-ranked, ambiguous-ranked, and hard-ranked---crossed with three seeds. Random train runs cross persistent random draws with the same three seeds. Each trained model is evaluated on SQuAD dev, AddSent, and AddOneSent, giving three evaluation commands per training run.}
\label{tab:run-matrix}

\end{minipage}
\vspace{12pt}
\noindent\begin{minipage}{\textwidth}

\centering\small\setlength{\tabcolsep}{9pt}
\begin{tabular}{lll}
\toprule
Model family & Checkpoint used & Approximate parameter count \\
\midrule
ELECTRA-small & \texttt{google/electra-small-discriminator} & 14M \\
BERT-base & \texttt{bert-base-uncased} & 110M \\
ELECTRA-base & \texttt{google/electra-base-discriminator} & 110M \\
RoBERTa-base & \texttt{roberta-base} & 125M \\
DistilBERT & \texttt{distilbert-base-uncased} & 67M \\
ALBERT-base & \texttt{albert-base-v2} & 11M--12M \\
DeBERTa-v3-base & \texttt{microsoft/deberta-v3-base} & 86M backbone; about 184M including embeddings \\
\bottomrule
\end{tabular}
\captionof{table}{Pretrained encoder checkpoints used in the multi-model panel. Counts are approximate and are included to make the compute/performance comparisons easier to interpret.}
\label{tab:checkpoints}

\end{minipage}
\vspace{12pt}
\noindent\begin{minipage}{\textwidth}

\centering\small\setlength{\tabcolsep}{7pt}
\begin{tabular}{llrrrrr}
\toprule
Model & GPU & Train cmds & Eval cmds & Train h & Eval h & Logged cmd h \\
\midrule
ELECTRA-small & 1$\times$ RTX 4090 & 42 & 126 & 2.76 & 2.27 & 5.04 \\
BERT-base & 1$\times$ RTX 4090 & 42 & 126 & 5.18 & 2.97 & 8.16 \\
ELECTRA-base & 1$\times$ RTX 4090 & 27 & 81 & 4.80 & 1.76 & 6.57 \\
RoBERTa-base & 1$\times$ RTX 4090 & 27 & 81 & 4.74 & 1.78 & 6.52 \\
DistilBERT & 1$\times$ RTX 4090 & 27 & 81 & 2.70 & 1.57 & 4.28 \\
ALBERT-base & 1$\times$ RTX 4090 & 27 & 81 & 6.49 & 1.80 & 8.30 \\
DeBERTa-v3-base & 1$\times$ A100 & 27 & 81 & 6.39 & 1.08 & 7.49 \\
\midrule
Total & -- & 219 & 657 & 33.06 & 13.23 & 46.34 \\
\bottomrule
\end{tabular}
\captionof{table}{Logged computational budget for the canonical experiment set. ``Logged cmd h'' sums command-log elapsed times and includes training, evaluation, dynamics aggregation, subset selection, and small materialization/audit commands. The Hugging Face Trainer \texttt{train\_runtime} fields sum to 29.52 training hours; we use command-log time as the primary budget because it includes command overhead. The total RunPod cost for the experiment suite was approximately \$70.}
\label{tab:compute-budget}

\end{minipage}

\clearpage
\twocolumn
\linenumbers

\section{Responsible NLP Checklist Details}
\label{app:responsible}
\subsection{Data, artifacts, and intended use}
This work uses existing public benchmark datasets and pretrained models. The source training data are SQuAD v1.1 examples, and robustness evaluation uses AddSent/AddOneSent-style adversarial examples derived from SQuAD. Our derived artifacts---subset files, cartography scores, metrics, figures, and scripts---are intended for research reproduction and analysis, not deployment.

\subsection{Human subjects and privacy}
We did not recruit human participants, collect new annotations, or interact with crowdworkers. We do not infer demographic attributes, perform user profiling, or process private user data. The datasets are public NLP benchmarks; their passages may contain named entities from the underlying Wikipedia text, but our study does not add personal data or attempt re-identification. We therefore view the direct privacy risk as low.

\subsection{Risks and scope}
The main risk is overgeneralization. Our results concern clean-source cartographic subset selection for SQuAD v1.1 evaluated on AddSent/AddOneSent adversarial distractors. They should not be interpreted as evidence that the evaluated models are robust in open-domain, multi-hop, long-context, unanswerable, generative, or safety-critical QA settings. The evaluated pretrained encoders may also inherit biases from their pretraining data; our experiments measure span-selection robustness under controlled distractor perturbations, not fairness or toxicity behavior. We mitigate these risks by limiting the claims to extractive adversarial QA under the evaluated benchmarks and by reporting failure cases, limitations, run variability, and model-specific exceptions.

\subsection{Environmental impact}
The study fine-tunes existing pretrained encoders rather than pretraining new language models. The canonical experiments used approximately 46.3 logged single-GPU command hours. We report hardware and GPU-hour estimates rather than a carbon-emissions estimate because we do not have enough information about provider-level power draw, utilization, datacenter energy mix, or regional carbon intensity to report such a number reliably.

\subsection{AI-assistant use}
AI assistance was used for drafting and editing support. The author inspected and controlled the manuscript claims, code, experimental results, and final text.

\clearpage
\nolinenumbers
\twocolumn[{%
\noindent\begin{minipage}{\textwidth}

\centering\footnotesize\setlength{\tabcolsep}{4pt}
\begin{tabular}{p{.17\linewidth}p{.19\linewidth}p{.25\linewidth}p{.30\linewidth}}
\toprule
Artifact & Source / citation & License or terms & Use in this work \\
\midrule
SQuAD v1.1 & \citet{squad} & CC BY-SA 4.0 & Clean-source training data and source examples for cartography. The grouped training map contains 87,599 examples. The benchmark is English reading comprehension over Wikipedia passages. \\
AddSent/AddOneSent & \citet{jia-adversarial} & SQuAD-derived challenge data; original adversarial-squad code released under MIT & Robustness evaluation. We report adversarial-only rows and paired original/adversarial failure rates rather than blending clean and adversarial rows. \\
Pretrained encoders & Model papers cited in Section~\ref{sec:training-eval} & Model-card licenses: Apache-2.0 for BERT, ELECTRA, DistilBERT, and ALBERT; MIT for RoBERTa and DeBERTa-v3 & Initializations for extractive QA fine-tuning. We do not release modified model weights in the reproducibility package. \\
Experiment package & This work & Research artifact & Contains run configurations, environment definitions, command logs, metrics, subset files, raw source-run dynamics, and figure/table-generation scripts. Intended for research reproduction and audit, not deployment. \\
\bottomrule
\end{tabular}
\captionof{table}{Scientific artifacts used or produced by the study. The paper cites the dataset, challenge-set, and model creators in the main text; this table adds license and intended-use details relevant to responsible artifact reporting.}
\label{tab:scientific-artifacts}

\end{minipage}
\vspace{18pt}
}]
\linenumbers

\section{Additional Method Details}
\label{app:methods}
\subsection{Training-dynamics aggregation}
For extractive QA, the model predicts start and end positions rather than a single class label. We therefore use the joint gold-span probability as the primary confidence field. For each SQuAD example id, confidence is the mean joint gold-span probability across retained training-dynamics records, variability is the standard deviation of that value, and correctness is the mean exact-match correctness across records. Because most grouped SQuAD examples are observed once per epoch in a three-epoch dynamics run, correctness is relatively coarse, while confidence and variability remain continuous.

\subsection{Ranked subset construction}
The ranked source subsets are not mutually exclusive geometric partitions or correctness categories. Easy-ranked examples are selected globally by high confidence, ambiguous-ranked examples globally by high variability, and hard-ranked examples globally by low confidence. Each ranked subset contains 33\% of the SQuAD training set. For visualization, we draw one fixed 10K sample per encoder and highlight only those sampled examples whose IDs belong to the corresponding global training subset; no within-sample re-ranking is performed. Persistent random subsets are materialized once per draw and then crossed with training seeds, so random 33\% performance is treated as a distribution over both subset draw and seed.

\subsection{Evaluation filtering and paired robustness}
The AddSent/AddOneSent-style files contain both original and adversarial rows. Our primary robustness metrics filter to adversarial-only rows. The paired analysis separately asks whether a model that answers an original row correctly fails after the corresponding adversarial perturbation. This distinction avoids mixing clean and adversarial examples in the primary metric while preserving a question-level perturbation view.

\section{Aggregate Diagnostics and Full Result Tables}
\label{app:aggregate}
This section contains aggregate tables and master figures only. Model-specific maps, global-subset intersection overlays, delta plots, and result tables are grouped by encoder family in Appendix~\ref{app:model-profiles}; they are not duplicated in this aggregate section. Tables~\ref{tab:map-stats}--\ref{tab:clean-squad} report the dataset, subset, and result audits; Figures~\ref{fig:addsent-amb-rand}--\ref{fig:all-model-dists} provide the aggregate exact-match and data-map diagnostics.

\clearpage

\nolinenumbers
\onecolumn
\noindent\begin{minipage}{\textwidth}

\centering
\small
\setlength{\tabcolsep}{5pt}
\begin{tabular}{lrrrrrr}
\toprule
Model & Mean conf. & Mean var. & Mean corr. & $=100\%$ & $\geq 50\%, <100\%$ & $<50\%$ \\
\midrule
ELECTRA-small & 0.367 & 0.121 & 0.531 & 30,735 & 17,371 & 39,493 \\
BERT-base & 0.510 & 0.157 & 0.639 & 38,793 & 19,234 & 29,572 \\
ELECTRA-base & 0.552 & 0.134 & 0.675 & 44,601 & 16,502 & 26,496 \\
RoBERTa-base & 0.537 & 0.128 & 0.664 & 44,024 & 15,873 & 27,702 \\
DistilBERT & 0.450 & 0.152 & 0.590 & 34,026 & 19,566 & 34,007 \\
ALBERT-base & 0.538 & 0.121 & 0.666 & 45,107 & 14,956 & 27,536 \\
DeBERTa-v3-base & 0.573 & 0.117 & 0.690 & 47,529 & 14,602 & 25,468 \\
\bottomrule
\end{tabular}
\captionof{table}{Summary statistics for the seven model-specific SQuAD data maps, each built from the model's source seed-42 full-data training dynamics. Each map contains 87,599 grouped SQuAD training examples. The final three columns count examples per correctness category.}
\label{tab:map-stats}

\end{minipage}

\vspace{3pt}
\noindent\begin{minipage}{\textwidth}

\centering
\small
\setlength{\tabcolsep}{3pt}
\renewcommand{\arraystretch}{1.12}
\begin{tabular}{lrrrrrr}
\toprule
Model & Easy $\cap$ ($=$100\%) & Amb. $\cap$ ($\geq$50\%, $<$100\%) & Hard $\cap$ ($<$50\%) & Easy$\cap$Amb. & Easy$\cap$Hard & Amb.$\cap$Hard \\
\midrule
ELECTRA-small & 85.3 & 42.8 & 97.1 & 44.1 & 0.0 & 1.7 \\
BERT-base & 96.9 & 49.5 & 88.9 & 15.7 & 0.0 & 18.5 \\
ELECTRA-base & 99.8 & 44.2 & 84.2 & 9.6 & 0.0 & 26.6 \\
RoBERTa-base & 99.4 & 42.4 & 86.5 & 12.3 & 0.0 & 23.9 \\
DistilBERT & 90.3 & 49.3 & 93.5 & 27.6 & 0.0 & 8.0 \\
ALBERT-base & 99.6 & 40.5 & 86.0 & 12.5 & 0.0 & 24.8 \\
DeBERTa-v3-base & 100.0 & 40.6 & 82.7 & 6.5 & 0.0 & 30.8 \\
\bottomrule
\end{tabular}
\captionof{table}{Model-specific ranked-subset audit. The first three columns give percentages of each globally ranked 33\% training subset whose correctness falls in the indicated category. The final columns give pairwise subset overlap as a percentage of subset size. Ranked selection rules may overlap.}
\label{tab:subset-overlap}

\end{minipage}

\vspace{3pt}
\noindent\begin{minipage}{\textwidth}

\centering
\small
\begin{tabular}{lrrr}
\toprule
Eval set & All rows & Original & Adversarial \\
\midrule
AddSent & 3560 & 1000 & 2560 \\
AddOneSent & 1787 & 1000 & 787 \\
\bottomrule
\end{tabular}
\captionof{table}{Composition of the AddSent/AddOneSent evaluation files. We report adversarial-only metrics as primary robustness results and use paired original/adversarial rows for the paired-failure analysis.}
\label{tab:split-audit}

\end{minipage}

\vspace{3pt}
\noindent\begin{minipage}{\textwidth}

\centering\small\setlength{\tabcolsep}{6pt}
\begin{tabular}{lrrrrr}
\toprule
Model & Map examples & 10K sample & Global subset size & Easy shown & Amb./Hard shown \\
\midrule
ELECTRA-small & 87,599 & 10,000 & 29,170 & 3,280 & 3,355/3,348 \\
BERT-base & 87,599 & 10,000 & 29,170 & 3,309 & 3,304/3,338 \\
ELECTRA-base & 87,599 & 10,000 & 29,170 & 3,248 & 3,365/3,379 \\
RoBERTa-base & 87,599 & 10,000 & 29,170 & 3,332 & 3,284/3,349 \\
DistilBERT & 87,599 & 10,000 & 29,170 & 3,307 & 3,308/3,380 \\
ALBERT-base & 87,599 & 10,000 & 29,170 & 3,322 & 3,329/3,331 \\
DeBERTa-v3-base & 87,599 & 10,000 & 29,170 & 3,263 & 3,381/3,320 \\
\bottomrule
\end{tabular}
\captionof{table}{Global-subset intersection audit for the overlay figures. Each panel shares a fixed 10K map sample (random state 1009); highlights are IDs also present in the globally selected 33\% subset. The counts vary because the subsets were ranked on the full training map rather than separately within the sample.}
\label{tab:sample-first-audit}

\end{minipage}

\vspace{3pt}
\noindent\begin{minipage}{\textwidth}

\centering
\small
\setlength{\tabcolsep}{4pt}
\renewcommand{\arraystretch}{1.15}
\begin{tabular}{lrrrrrr}
\toprule
Model & Full & Random & Easy & Ambiguous & Hard & $\Delta_{Amb-Rand}$ \\
\midrule
ELECTRA-small & \textbf{42.60 $\pm$ 1.23} & 34.08 $\pm$ 1.06 & 29.78 $\pm$ 0.80 & 34.22 $\pm$ 0.85 & 29.70 $\pm$ 0.74 & +0.13 \\
BERT-base & \textbf{51.00 $\pm$ 0.23} & 44.30 $\pm$ 1.28 & 38.53 $\pm$ 1.00 & 47.42 $\pm$ 1.35 & 38.37 $\pm$ 0.62 & +3.12 \\
ELECTRA-base & \textbf{62.57 $\pm$ 0.78} & 54.85 $\pm$ 1.60 & 49.45 $\pm$ 1.90 & 59.45 $\pm$ 1.39 & 48.49 $\pm$ 0.09 & +4.61 \\
RoBERTa-base & \textbf{61.25 $\pm$ 0.52} & 57.76 $\pm$ 1.45 & 50.61 $\pm$ 0.99 & 59.90 $\pm$ 0.79 & 49.27 $\pm$ 1.25 & +2.14 \\
DistilBERT & \textbf{46.08 $\pm$ 1.14} & 38.34 $\pm$ 1.41 & 32.30 $\pm$ 0.57 & 40.08 $\pm$ 1.22 & 35.26 $\pm$ 1.26 & +1.74 \\
ALBERT-base & \textbf{55.40 $\pm$ 1.04} & 49.77 $\pm$ 1.50 & 41.84 $\pm$ 1.40 & 52.90 $\pm$ 2.06 & 43.53 $\pm$ 1.73 & +3.14 \\
DeBERTa-v3-base & 74.67 $\pm$ 0.90 & 71.84 $\pm$ 0.94 & 60.51 $\pm$ 2.19 & \textbf{74.99 $\pm$ 1.11} & 59.99 $\pm$ 0.26 & +3.14 \\
\bottomrule
\end{tabular}
\captionof{table}{Adversarial-only AddSent exact-match accuracy. Values are means $\pm$ standard deviations over training seeds and random subset draws where applicable; the best condition in each row is bolded.}
\label{tab:addsent-em}

\end{minipage}

\clearpage
\noindent\begin{minipage}{\textwidth}

\centering
\small
\setlength{\tabcolsep}{4pt}
\renewcommand{\arraystretch}{1.15}
\begin{tabular}{lrrrrrr}
\toprule
Model & Full & Random & Easy & Ambiguous & Hard & $\Delta_{Amb-Rand}$ \\
\midrule
ELECTRA-small & \textbf{42.57 $\pm$ 1.16} & 33.81 $\pm$ 1.16 & 29.56 $\pm$ 1.18 & 34.43 $\pm$ 1.13 & 28.29 $\pm$ 0.95 & +0.62 \\
BERT-base & \textbf{50.49 $\pm$ 0.60} & 43.91 $\pm$ 1.34 & 39.35 $\pm$ 1.51 & 47.73 $\pm$ 2.04 & 35.96 $\pm$ 0.77 & +3.82 \\
ELECTRA-base & \textbf{62.35 $\pm$ 0.32} & 54.36 $\pm$ 1.92 & 50.28 $\pm$ 2.17 & 59.81 $\pm$ 1.20 & 47.27 $\pm$ 0.55 & +5.45 \\
RoBERTa-base & \textbf{62.13 $\pm$ 0.13} & 57.81 $\pm$ 1.31 & 51.29 $\pm$ 0.60 & 59.76 $\pm$ 0.57 & 48.84 $\pm$ 1.41 & +1.95 \\
DistilBERT & \textbf{45.70 $\pm$ 1.48} & 37.78 $\pm$ 1.27 & 32.70 $\pm$ 0.73 & 40.24 $\pm$ 1.27 & 33.55 $\pm$ 1.44 & +2.46 \\
ALBERT-base & \textbf{55.65 $\pm$ 0.92} & 49.99 $\pm$ 1.45 & 41.00 $\pm$ 1.73 & 53.66 $\pm$ 1.73 & 43.03 $\pm$ 1.80 & +3.68 \\
DeBERTa-v3-base & 75.01 $\pm$ 0.73 & 71.83 $\pm$ 1.05 & 61.33 $\pm$ 2.27 & \textbf{75.35 $\pm$ 1.44} & 58.20 $\pm$ 0.77 & +3.52 \\
\bottomrule
\end{tabular}
\captionof{table}{Adversarial-only AddOneSent exact-match accuracy. Values are means $\pm$ standard deviations over training seeds and random subset draws where applicable; the best condition in each row is bolded.}
\label{tab:addonesent-em}

\end{minipage}

\vspace{8pt}
\noindent\begin{minipage}{\textwidth}

\centering
\small
\setlength{\tabcolsep}{4pt}
\renewcommand{\arraystretch}{1.15}
\begin{tabular}{lrrrrrr}
\toprule
Model & Full & Random & Easy & Ambiguous & Hard & $\Delta_{Amb-Rand}$ \\
\midrule
ELECTRA-small & \textbf{49.37 $\pm$ 1.41} & 41.08 $\pm$ 1.09 & 34.90 $\pm$ 0.86 & 39.82 $\pm$ 0.82 & 40.36 $\pm$ 0.80 & -1.26 \\
BERT-base & \textbf{57.66 $\pm$ 0.39} & 51.27 $\pm$ 1.32 & 43.70 $\pm$ 1.06 & 53.45 $\pm$ 1.41 & 49.98 $\pm$ 1.04 & +2.18 \\
ELECTRA-base & \textbf{68.37 $\pm$ 1.21} & 61.25 $\pm$ 1.70 & 53.78 $\pm$ 1.80 & 65.27 $\pm$ 1.53 & 60.15 $\pm$ 0.35 & +4.02 \\
RoBERTa-base & \textbf{67.67 $\pm$ 0.26} & 64.14 $\pm$ 1.26 & 54.95 $\pm$ 1.02 & 65.58 $\pm$ 1.02 & 60.21 $\pm$ 1.57 & +1.44 \\
DistilBERT & \textbf{53.15 $\pm$ 1.17} & 45.61 $\pm$ 1.33 & 37.64 $\pm$ 0.90 & 46.31 $\pm$ 1.27 & 47.02 $\pm$ 1.39 & +0.71 \\
ALBERT-base & \textbf{61.69 $\pm$ 0.78} & 56.03 $\pm$ 1.50 & 46.56 $\pm$ 1.48 & 58.05 $\pm$ 1.98 & 53.94 $\pm$ 1.45 & +2.02 \\
DeBERTa-v3-base & 80.03 $\pm$ 1.14 & 78.15 $\pm$ 0.83 & 64.85 $\pm$ 2.24 & \textbf{80.33 $\pm$ 1.29} & 73.35 $\pm$ 0.29 & +2.18 \\
\bottomrule
\end{tabular}
\captionof{table}{Adversarial-only AddSent token F1. Values are means $\pm$ standard deviations over training seeds and random subset draws where applicable; the best condition in each row is bolded.}
\label{tab:addsent-f1}

\end{minipage}

\vspace{8pt}
\noindent\begin{minipage}{\textwidth}

\centering
\small
\setlength{\tabcolsep}{4pt}
\renewcommand{\arraystretch}{1.15}
\begin{tabular}{lrrrrrr}
\toprule
Model & Full & Random & Easy & Ambiguous & Hard & $\Delta_{Amb-Rand}$ \\
\midrule
ELECTRA-small & \textbf{49.10 $\pm$ 1.39} & 40.92 $\pm$ 1.13 & 34.55 $\pm$ 1.07 & 39.89 $\pm$ 1.01 & 39.43 $\pm$ 0.91 & -1.04 \\
BERT-base & \textbf{57.10 $\pm$ 0.99} & 50.47 $\pm$ 1.40 & 44.19 $\pm$ 1.66 & 53.50 $\pm$ 2.14 & 47.47 $\pm$ 1.15 & +3.03 \\
ELECTRA-base & \textbf{68.45 $\pm$ 0.71} & 60.57 $\pm$ 2.18 & 54.22 $\pm$ 2.19 & 65.47 $\pm$ 1.53 & 59.07 $\pm$ 0.91 & +4.90 \\
RoBERTa-base & \textbf{68.23 $\pm$ 0.09} & 63.95 $\pm$ 1.30 & 55.42 $\pm$ 0.36 & 65.33 $\pm$ 0.74 & 59.82 $\pm$ 1.53 & +1.38 \\
DistilBERT & \textbf{52.50 $\pm$ 1.54} & 44.80 $\pm$ 1.23 & 38.24 $\pm$ 1.20 & 46.37 $\pm$ 1.58 & 45.29 $\pm$ 1.60 & +1.56 \\
ALBERT-base & \textbf{62.00 $\pm$ 0.90} & 55.99 $\pm$ 1.48 & 45.62 $\pm$ 1.81 & 58.25 $\pm$ 1.77 & 53.53 $\pm$ 1.79 & +2.26 \\
DeBERTa-v3-base & 80.11 $\pm$ 0.80 & 77.84 $\pm$ 1.09 & 65.60 $\pm$ 2.36 & \textbf{80.35 $\pm$ 1.65} & 72.11 $\pm$ 1.04 & +2.52 \\
\bottomrule
\end{tabular}
\captionof{table}{Adversarial-only AddOneSent token F1. Values are means $\pm$ standard deviations over training seeds and random subset draws where applicable; the best condition in each row is bolded.}
\label{tab:addonesent-f1}

\end{minipage}

\vspace{8pt}
\noindent\begin{minipage}{\textwidth}

\centering
\small
\setlength{\tabcolsep}{5pt}
\renewcommand{\arraystretch}{1.15}
\begin{tabular}{lrrrrrr}
\toprule
 & \multicolumn{3}{c}{AddSent failure rate} & \multicolumn{3}{c}{AddOneSent failure rate} \\
\cmidrule(lr){2-4}\cmidrule(lr){5-7}
Model & Rand & Amb. & $\Delta$ & Rand & Amb. & $\Delta$ \\
\midrule
ELECTRA-small & 68.0 & \textbf{67.9} & -0.1 & \textbf{53.3} & 54.0 & +0.7 \\
BERT-base & 59.8 & \textbf{57.6} & -2.3 & 43.5 & \textbf{40.5} & -3.0 \\
ELECTRA-base & 49.6 & \textbf{44.9} & -4.8 & 36.3 & \textbf{31.6} & -4.7 \\
RoBERTa-base & 43.9 & \textbf{40.5} & -3.3 & 31.7 & \textbf{30.4} & -1.3 \\
DistilBERT & 64.7 & \textbf{64.2} & -0.5 & 47.4 & \textbf{46.2} & -1.2 \\
ALBERT-base & 53.8 & \textbf{50.7} & -3.1 & 39.6 & \textbf{36.1} & -3.5 \\
DeBERTa-v3-base & 26.3 & \textbf{22.3} & -4.0 & 18.2 & \textbf{15.2} & -3.0 \\
\bottomrule
\end{tabular}
\captionof{table}{Paired robustness failure rates. A failure is counted when a model answers the original row correctly but fails on at least one adversarial variant of the same base question. Lower is better, so the lower value between random and ambiguous is bolded.}
\label{tab:paired-failure}

\end{minipage}

\vspace{8pt}

\clearpage
\begingroup

\centering\includegraphics[width=.96\linewidth]{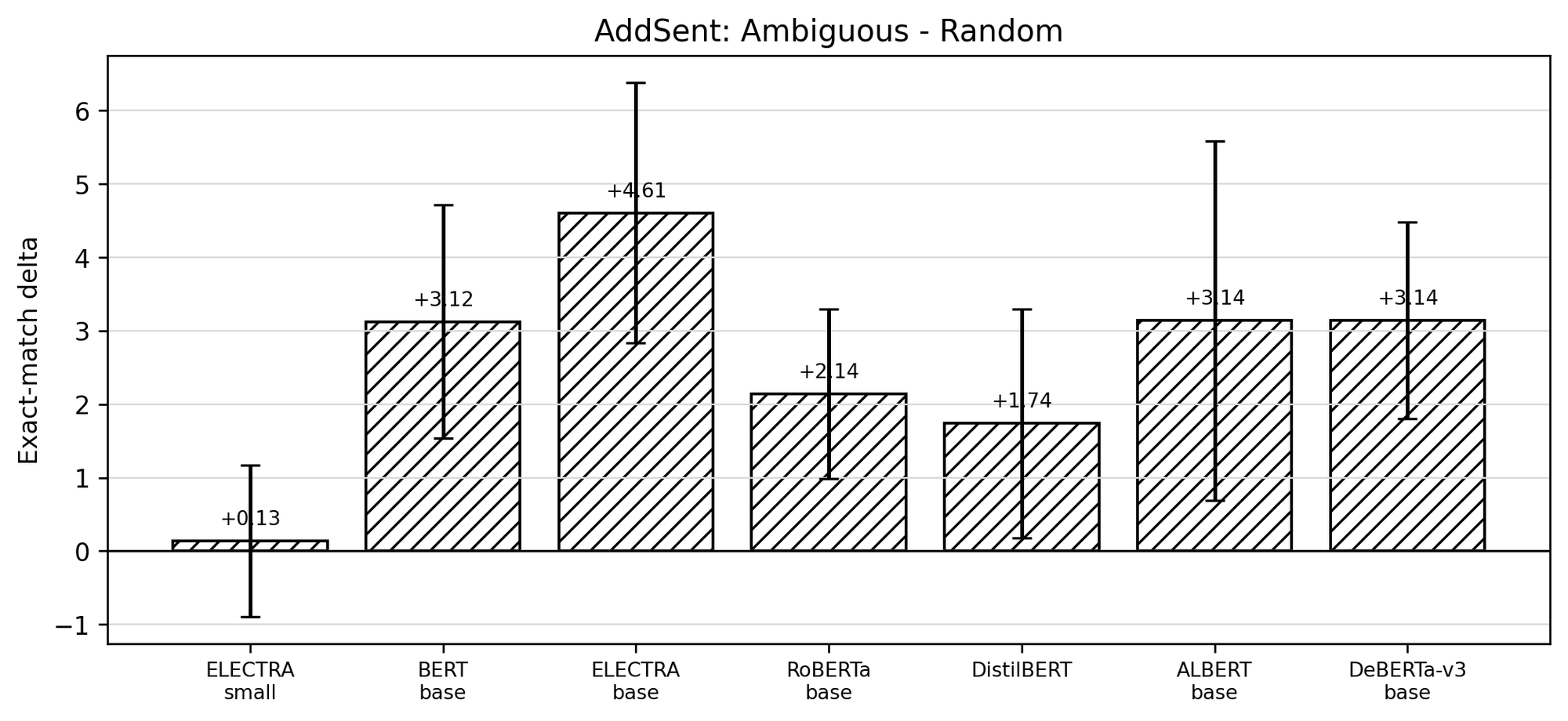}
\captionof{figure}{Adversarial-only AddSent exact-match deltas for ambiguous-ranked 33\% training minus random 33\% training. Error bars are approximate 95\% confidence intervals from run-level standard errors.}
\label{fig:addsent-amb-rand}

\par\endgroup

\vspace{14pt}
\begingroup

\centering\includegraphics[width=.96\linewidth]{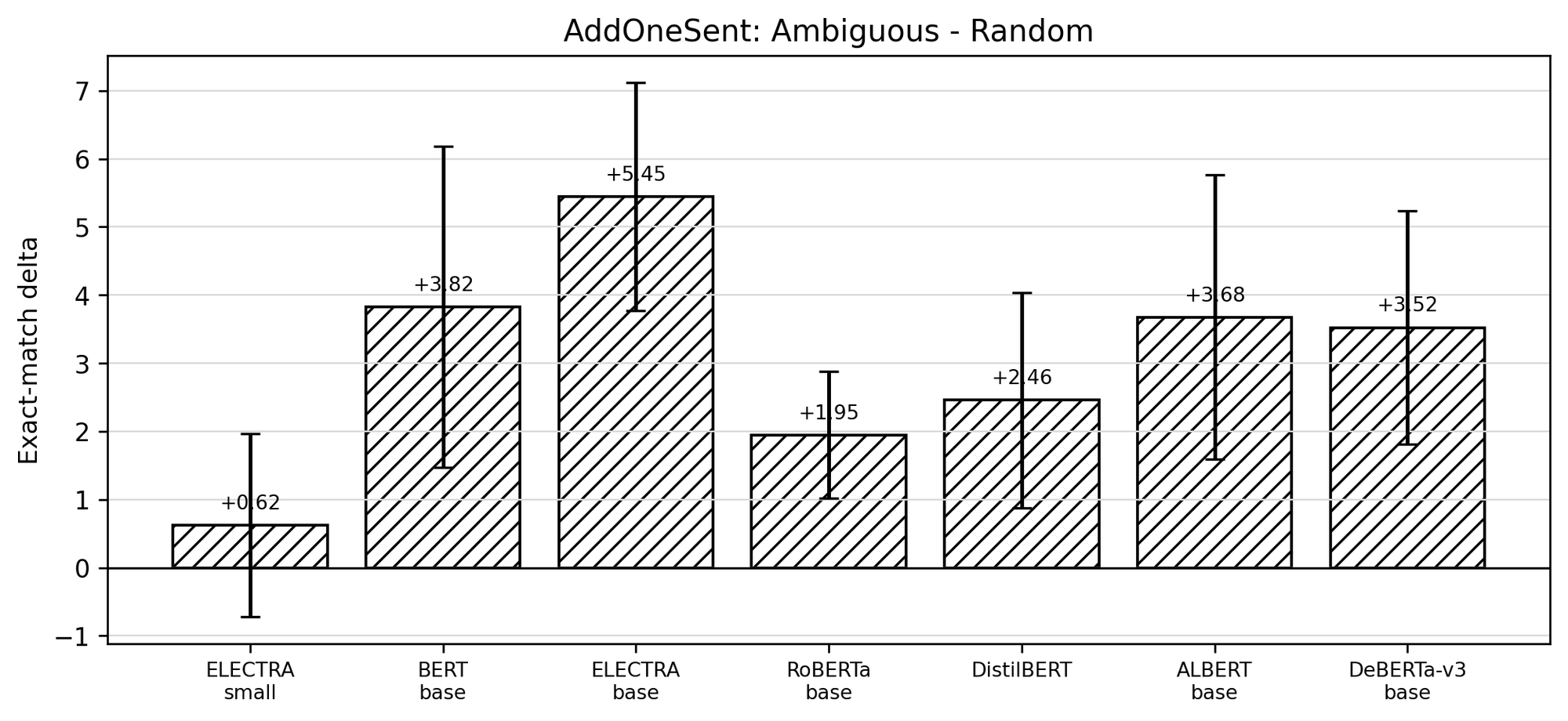}
\captionof{figure}{Adversarial-only AddOneSent exact-match deltas for ambiguous-ranked 33\% training minus random 33\% training. Error bars are approximate 95\% confidence intervals from run-level standard errors.}
\label{fig:addonesent-amb-rand}

\par\endgroup

\vspace{6pt}
\noindent\begin{minipage}{\textwidth}
\centering
\footnotesize
\setlength{\tabcolsep}{4pt}
\renewcommand{\arraystretch}{1.05}
\begin{tabular}{lrrrrrr}
\toprule
 & \multicolumn{3}{c}{SQuAD-dev EM} & \multicolumn{3}{c}{SQuAD-dev F1} \\
\cmidrule(lr){2-4}\cmidrule(lr){5-7}
Model & Full & Random & Ambiguous & Full & Random & Ambiguous \\
\midrule
ELECTRA-small & \textbf{77.48 $\pm$ 0.18} & 71.33 $\pm$ 0.39 & 72.41 $\pm$ 0.33 & \textbf{85.47 $\pm$ 0.15} & 80.75 $\pm$ 0.38 & 80.67 $\pm$ 0.30 \\
BERT-base & \textbf{80.65 $\pm$ 0.29} & 76.27 $\pm$ 0.20 & 78.76 $\pm$ 0.16 & \textbf{88.34 $\pm$ 0.17} & 85.07 $\pm$ 0.19 & 86.39 $\pm$ 0.12 \\
ELECTRA-base & \textbf{85.29 $\pm$ 0.07} & 82.77 $\pm$ 0.37 & 84.51 $\pm$ 0.19 & \textbf{91.95 $\pm$ 0.08} & 90.07 $\pm$ 0.24 & 91.17 $\pm$ 0.11 \\
RoBERTa-base & \textbf{85.39 $\pm$ 0.15} & 83.47 $\pm$ 0.25 & 85.20 $\pm$ 0.10 & \textbf{91.91 $\pm$ 0.18} & 90.49 $\pm$ 0.16 & 91.41 $\pm$ 0.13 \\
DistilBERT & \textbf{77.54 $\pm$ 0.12} & 71.77 $\pm$ 0.38 & 73.92 $\pm$ 0.31 & \textbf{85.77 $\pm$ 0.08} & 81.34 $\pm$ 0.32 & 82.25 $\pm$ 0.13 \\
ALBERT-base & \textbf{82.34 $\pm$ 0.34} & 80.03 $\pm$ 0.34 & 82.22 $\pm$ 0.25 & \textbf{89.83 $\pm$ 0.18} & 88.22 $\pm$ 0.26 & 89.49 $\pm$ 0.11 \\
DeBERTa-v3-base & \textbf{87.72 $\pm$ 0.21} & 85.99 $\pm$ 0.27 & 87.32 $\pm$ 0.09 & \textbf{93.62 $\pm$ 0.14} & 92.46 $\pm$ 0.17 & 93.26 $\pm$ 0.05 \\
\midrule
Mean & \textbf{82.34} & 78.80 & 80.62 & \textbf{89.56} & 86.91 & 87.80 \\
\bottomrule
\end{tabular}
\captionof{table}{Clean/original SQuAD-dev exact-match and F1 for the three central training conditions. Values are means $\pm$ standard deviations over training seeds and random subset draws where applicable; the best condition in each row is bolded per metric. The clean-domain ordering matches the adversarial-only results: ambiguous-ranked 33\% beats random 33\% on EM in 7 of 7 models and on F1 in 6 of 7 (ELECTRA-small is the exception), while full-data training is best in all models on both metrics.}
\label{tab:clean-squad}

\end{minipage}

\vspace{14pt}

\clearpage
\begingroup

\centering\includegraphics[width=.96\linewidth]{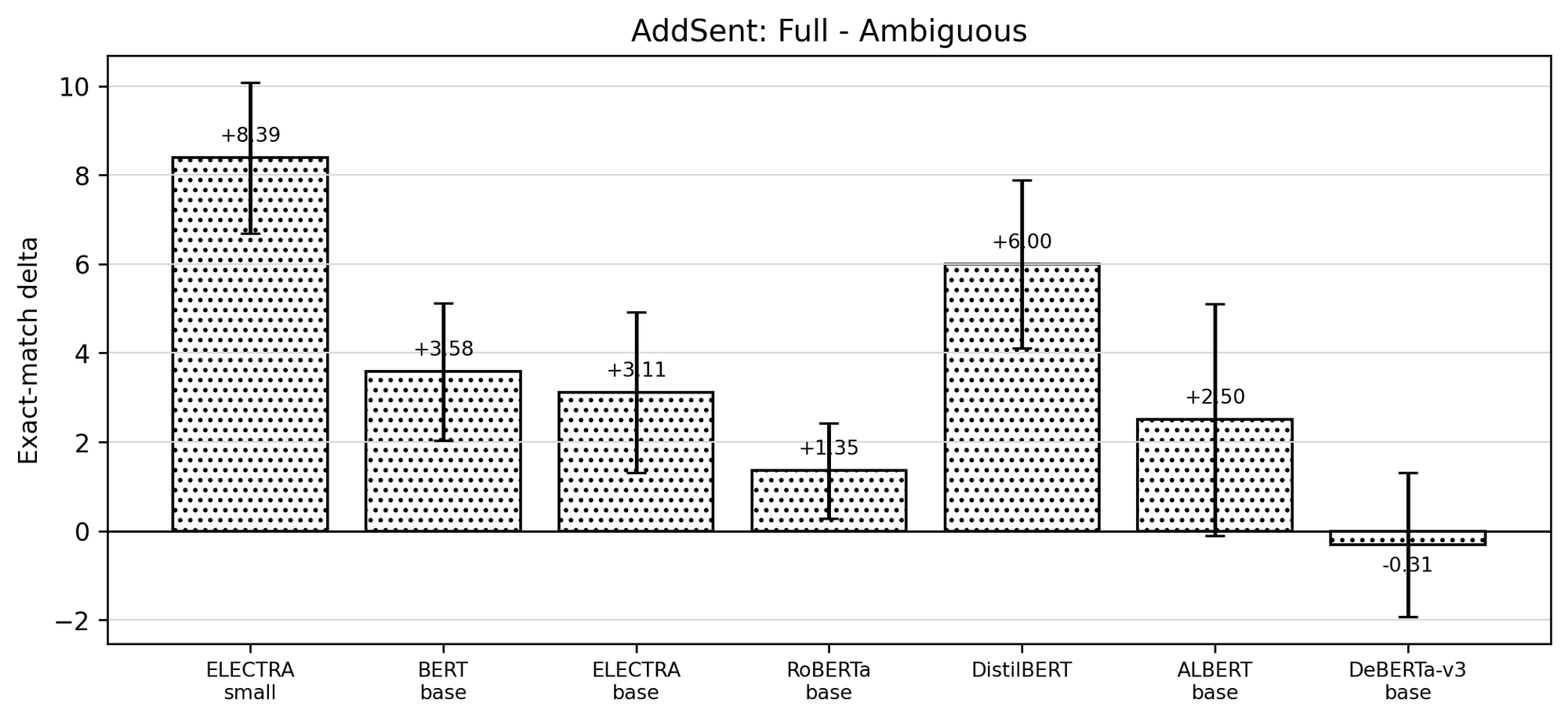}
\captionof{figure}{Adversarial-only AddSent exact-match deltas for full-data training minus ambiguous-ranked 33\% training. Error bars are approximate 95\% confidence intervals from run-level standard errors.}
\label{fig:addsent-full-amb}

\par\endgroup

\vspace{14pt}
\begingroup

\centering\includegraphics[width=.96\linewidth]{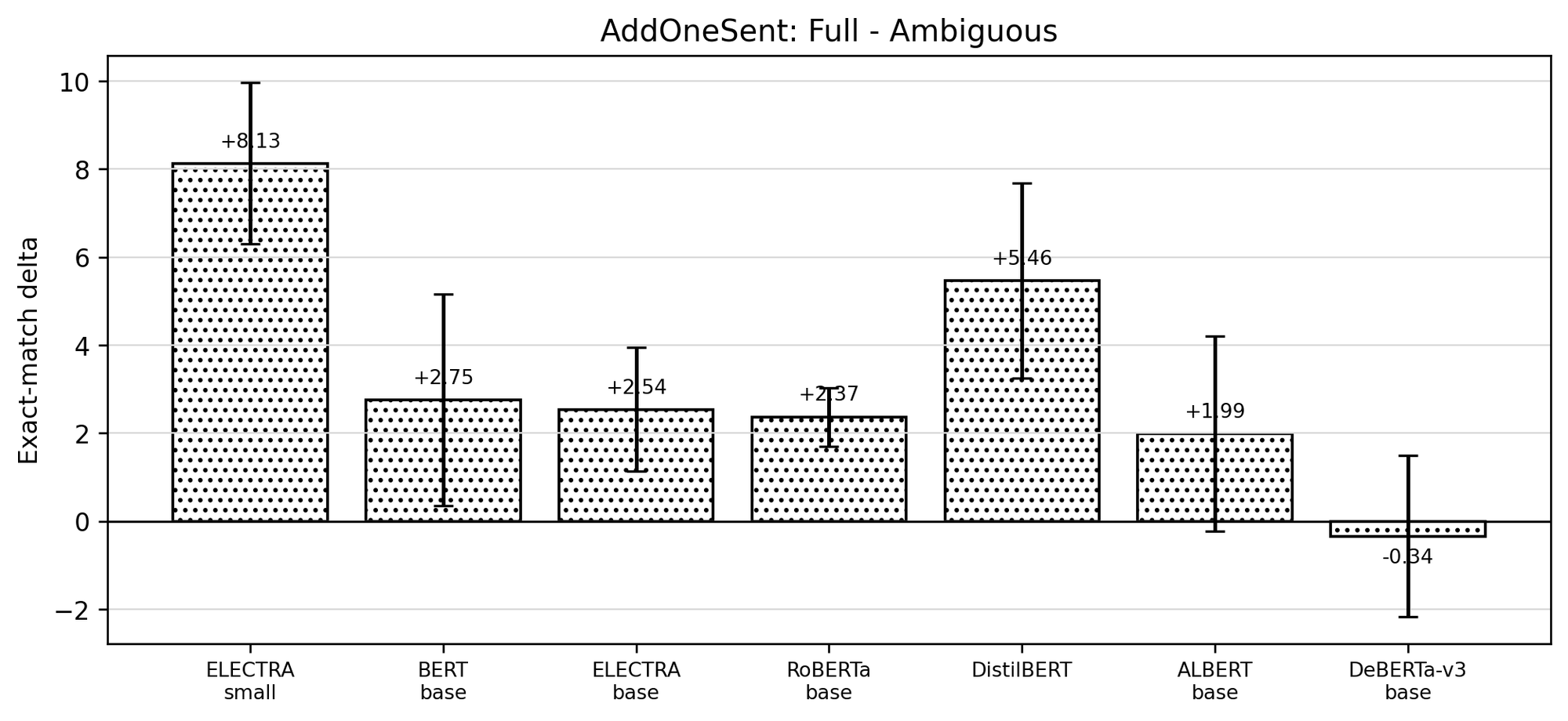}
\captionof{figure}{Adversarial-only AddOneSent exact-match deltas for full-data training minus ambiguous-ranked 33\% training. Error bars are approximate 95\% confidence intervals from run-level standard errors.}
\label{fig:addonesent-full-amb}

\par\endgroup

\vspace{14pt}

\clearpage
\begingroup

\centering\includegraphics[width=.89\linewidth]{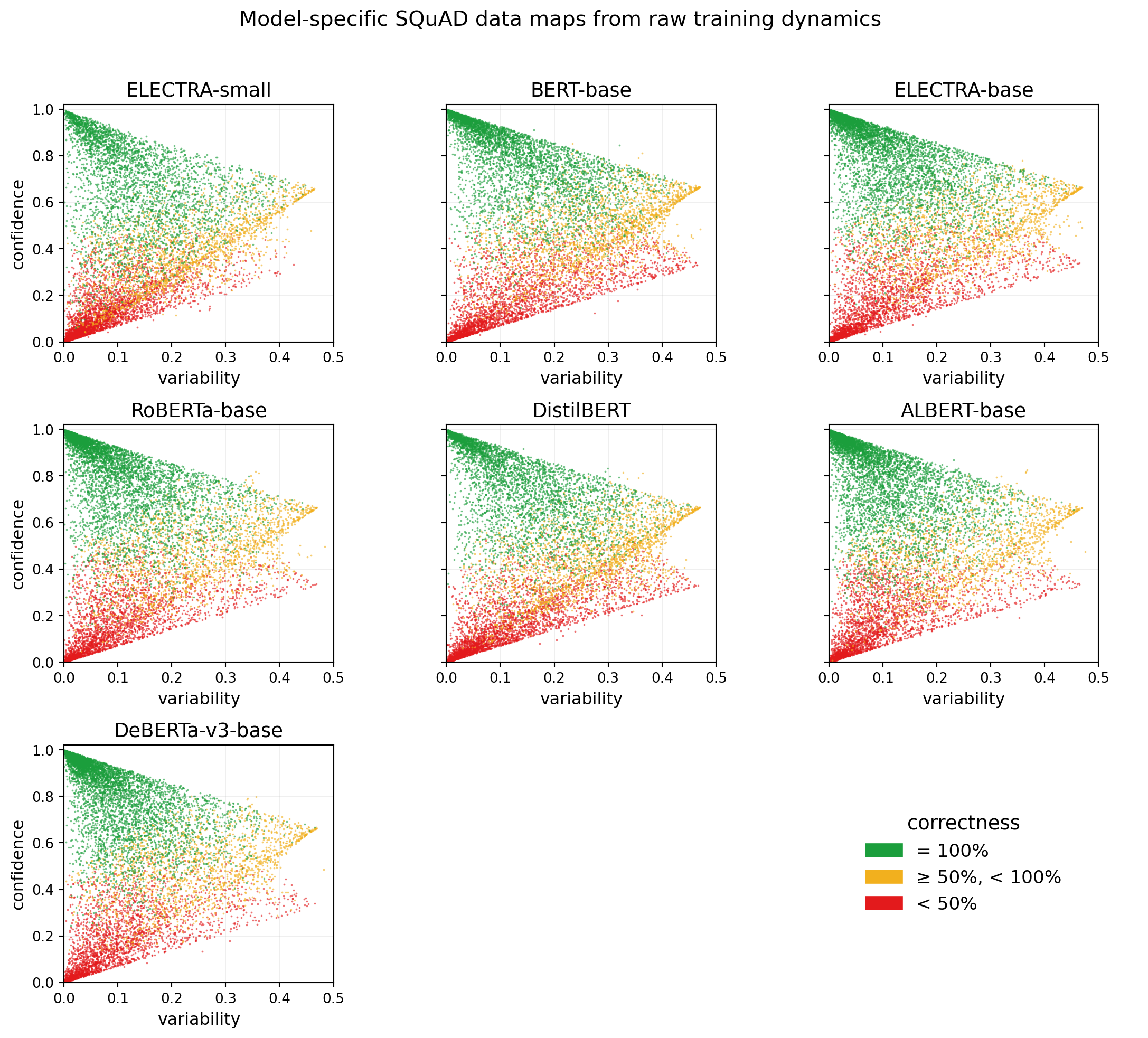}
\captionof{figure}{Model-specific SQuAD data maps for all seven encoder families. Each panel plots confidence against variability for examples sampled from the model's own clean-source training dynamics. Colors encode exact-match correctness using the same green/amber/red convention as Figure~\ref{fig:squad-datamap}. The maps show broadly similar cartographic structure across architectures while preserving model-specific differences in density and spread.}
\label{fig:all-model-datamaps}

\par\endgroup

\vspace{14pt}

\clearpage
\begingroup

\centering\includegraphics[height=.81\textheight]{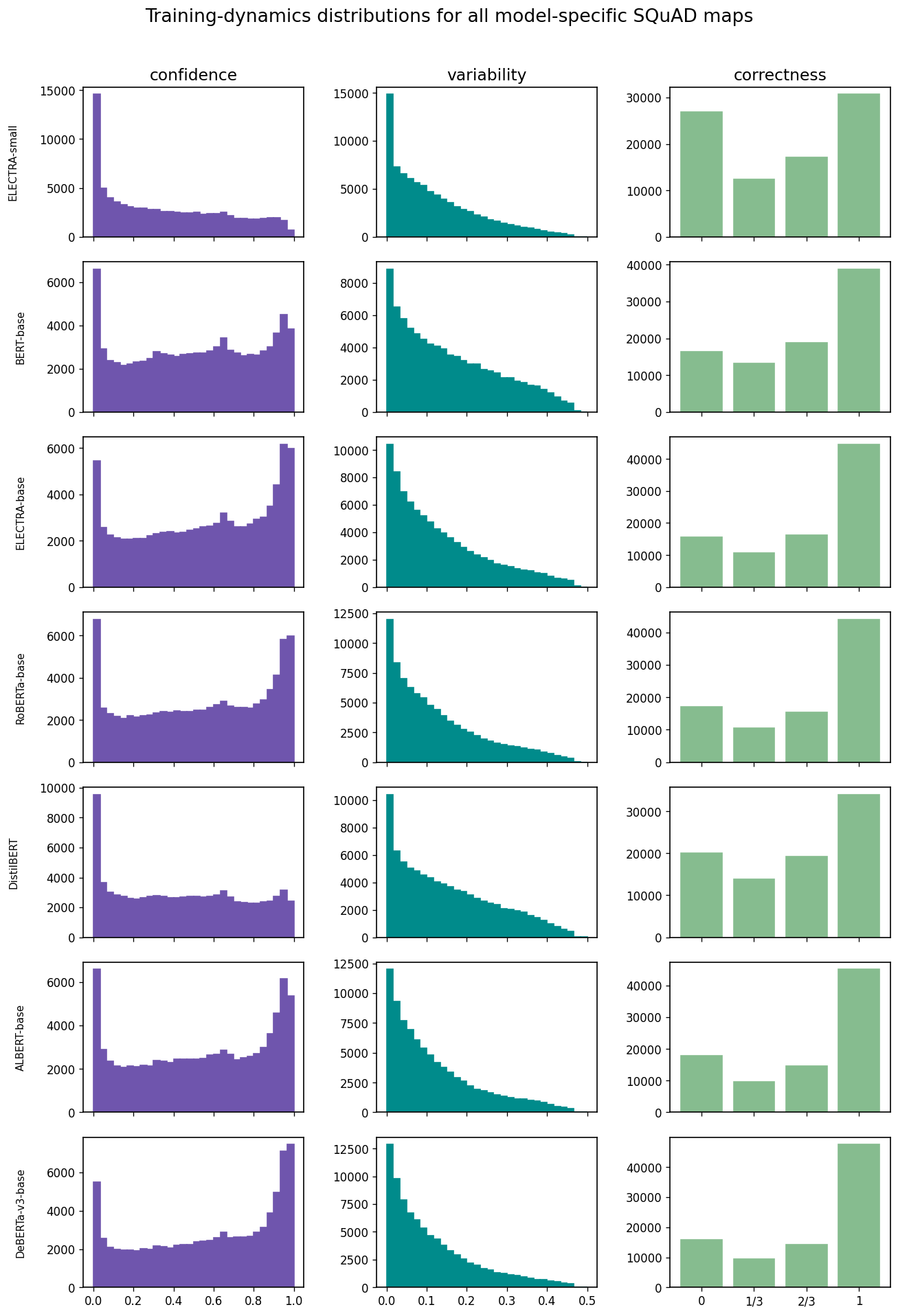}
\captionof{figure}{Confidence, variability, and correctness distributions for all model-specific SQuAD maps. Correctness is discrete because it records the fraction of dynamics records with exact-match-correct predictions.}
\label{fig:all-model-dists}

\par\endgroup

\vspace{14pt}

\clearpage
\noindent\begin{minipage}{\textwidth}

\raggedright\section{Qualitative Examples}\label{app:qualitative}
\centering\footnotesize\setlength{\tabcolsep}{2pt}\renewcommand{\arraystretch}{1.85}
\begin{tabular}{p{.15\linewidth}p{.11\linewidth}p{.33\linewidth}p{.13\linewidth}rrr}
\toprule
Correctness & Article & Question & Gold answer & Conf. & Var. & Corr. \\
\midrule
\colorbox{green!15}{$=$100\%} & Neptune & Who discovered Triton? & \colorbox{orange!18}{William Lassell} & 0.70 & 0.40 & 1.00 \\
 & Computer & When was the first analog computer in the form of a tide-predicting machine created? & \colorbox{orange!18}{1872} & 0.87 & 0.16 & 1.00 \\
\midrule
\colorbox{yellow!25}{$\geq$50\%, $<$100\%} & Hellenistic period & When did Cassander die? & \colorbox{orange!18}{298 BCE} & 0.42 & 0.42 & 0.50 \\
 & Beyoncé & Which year did Beyoncé and her father part business ways? & \colorbox{orange!18}{2010} & 0.65 & 0.46 & 0.67 \\
\midrule
\colorbox{red!12}{$<$50\%} & Olympic torch relay & After the torch bearer disappeared into a warehouse, on what road did the relay reappear? & \cellbg{orange!18}{Van Ness Avenue} & 0.23 & 0.33 & 0.33 \\
 & Galicia & Which empire turned Galicia into one of its provinces in the 3rd century AD? & \cellbg{orange!18}{Roman Empire} & 0.19 & 0.26 & 0.33 \\
\bottomrule
\end{tabular}
\captionof{table}{Illustrative SQuAD training examples from the ELECTRA-small source data map, grouped by correctness category. Gold answers appear in orange. We use nonzero-variability examples so that the table reflects movement over training rather than only stable corner cases.}
\label{tab:squad-qualitative}

\end{minipage}

\vspace{15pt}
\noindent\begin{minipage}{\textwidth}

\centering\footnotesize\setlength{\tabcolsep}{2pt}\renewcommand{\arraystretch}{1.22}
\begin{tabular}{p{.12\linewidth}p{.10\linewidth}p{.25\linewidth}p{.18\linewidth}p{.11\linewidth}p{.09\linewidth}}
\toprule
Run & Article & Question & Gold answer(s) & Original prediction & Adversarial prediction \\
\midrule
ELECTRA-small full & Ctenophora & What does the Beroe eat? & \cellbg{orange!18}{other ctenophores; ctenophores} & \cellbg{green!18}{other ctenophores} & \cellbg{red!12}{hamsters} \\
BERT ambiguous & Warsaw & Over how many species of trees can be found in the Saxon Garden? & \cellbg{orange!18}{100; over 100} & \cellbg{green!18}{100} & \cellbg{red!12}{600} \\
ELECTRA-base ambiguous & Nikola Tesla & What was Galileo Ferraris? & \cellbg{orange!18}{physicist; Italian physicist} & \cellbg{green!18}{Italian physicist} & \cellbg{red!12}{Rosetta Lamborghinis} \\
RoBERTa ambiguous & Normans & What kind of bureaucracy did the Normans institute? & \cellbg{orange!18}{meritocratic} & \cellbg{green!18}{meritocratic} & \cellbg{red!12}{deep} \\
DeBERTa ambiguous & Teacher & What objective would be labeled as practical? & \cellbg{orange!18}{skill} & \cellbg{green!18}{skill} & \cellbg{red!12}{hamster} \\
\bottomrule
\end{tabular}
\captionof{table}{Adversarial output examples from different SQuAD articles. Each row shows a case where the model is exact-match correct on the original row but changes to an incorrect answer after the AddSent distractor is inserted. The adversarial prediction is the model's prediction on the adversarially perturbed row. Orange marks gold answers, green marks the original correct output, and red marks the adversarial output selected from the distractor.}
\label{tab:adversarial-output-examples}

\end{minipage}

\vspace{15pt}

\clearpage
\begingroup

\noindent\begin{minipage}[t]{.495\linewidth}\raggedright\section{Model-Specific Profiles}\label{app:model-profiles}
This appendix groups the model-specific diagnostics by encoder family. Each profile provides the model-specific SQuAD data map, global training-subset intersections with a fixed 10K sample, the exact-match delta plot, and the full adversarial-only result table for that model.\par\end{minipage}\par\medskip
\raggedright\subsection{ELECTRA-small}
\centering
\includegraphics[width=.81\linewidth]{figures/fig01_electra_map.png}
\captionof{figure}{ELECTRA-small SQuAD data map and dynamics marginals from the fixed 10K sample.}
\label{fig:profile-map-electra_small}
\vspace{1em}
\includegraphics[width=.81\linewidth]{figures/fig03_electra_sample_first.png}
\captionof{figure}{ELECTRA-small global easy-, ambiguous-, and hard-ranked 33\% training subsets intersected with the same fixed 10K map sample. Gray points show the background sample; colored points are sampled members of the relevant global subset. Colors denote correctness categories and may be mixed within each panel. Highlight membership follows the globally ranked training subsets.}
\label{fig:profile-overlay-electra_small}

\par\endgroup

\clearpage
\noindent\begin{minipage}{\textwidth}

\centering
\small
\setlength{\tabcolsep}{5pt}
\begin{tabular}{lrrrr}
\toprule
Condition & AddSent EM & AddSent F1 & AddOneSent EM & AddOneSent F1 \\
\midrule
Full & \textbf{42.60 $\pm$ 1.23} & \textbf{49.37 $\pm$ 1.41} & \textbf{42.57 $\pm$ 1.16} & \textbf{49.10 $\pm$ 1.39} \\
Rand & 34.08 $\pm$ 1.06 & 41.08 $\pm$ 1.09 & 33.81 $\pm$ 1.16 & 40.92 $\pm$ 1.13 \\
Easy & 29.78 $\pm$ 0.80 & 34.90 $\pm$ 0.86 & 29.56 $\pm$ 1.18 & 34.55 $\pm$ 1.07 \\
Amb. & 34.22 $\pm$ 0.85 & 39.82 $\pm$ 0.82 & 34.43 $\pm$ 1.13 & 39.89 $\pm$ 1.01 \\
Hard & 29.70 $\pm$ 0.74 & 40.36 $\pm$ 0.80 & 28.29 $\pm$ 0.95 & 39.43 $\pm$ 0.91 \\
\bottomrule
\end{tabular}
\captionof{table}{Adversarial-only results for ELECTRA-small. Best value in each column is bolded.}
\label{tab:model-specific-electra-small}

\end{minipage}

\vspace{12pt}
\begingroup

\centering
\includegraphics[width=.72\linewidth]{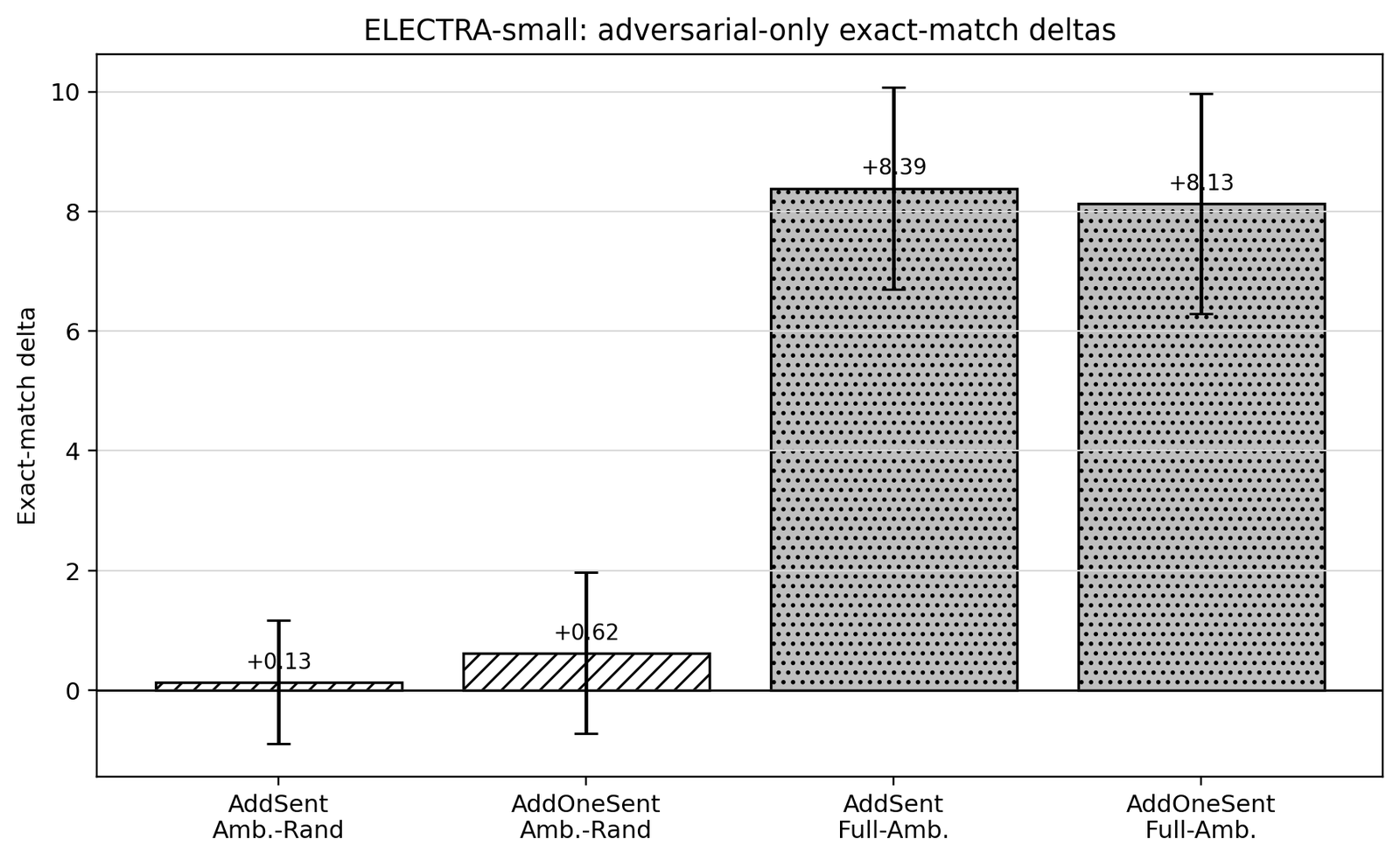}
\captionof{figure}{ELECTRA-small adversarial-only exact-match delta bars. Amb.-Rand bars compare ambiguous-ranked 33\% training against the random 33\% baseline; Full-Amb. bars compare full-data training against ambiguous-ranked 33\%. Error bars are approximate 95\% confidence intervals from run-level standard errors.}
\label{fig:profile-delta-electra_small}

\par\endgroup

\clearpage
\begingroup

\raggedright\subsection{BERT-base}
\centering
\includegraphics[width=.81\linewidth]{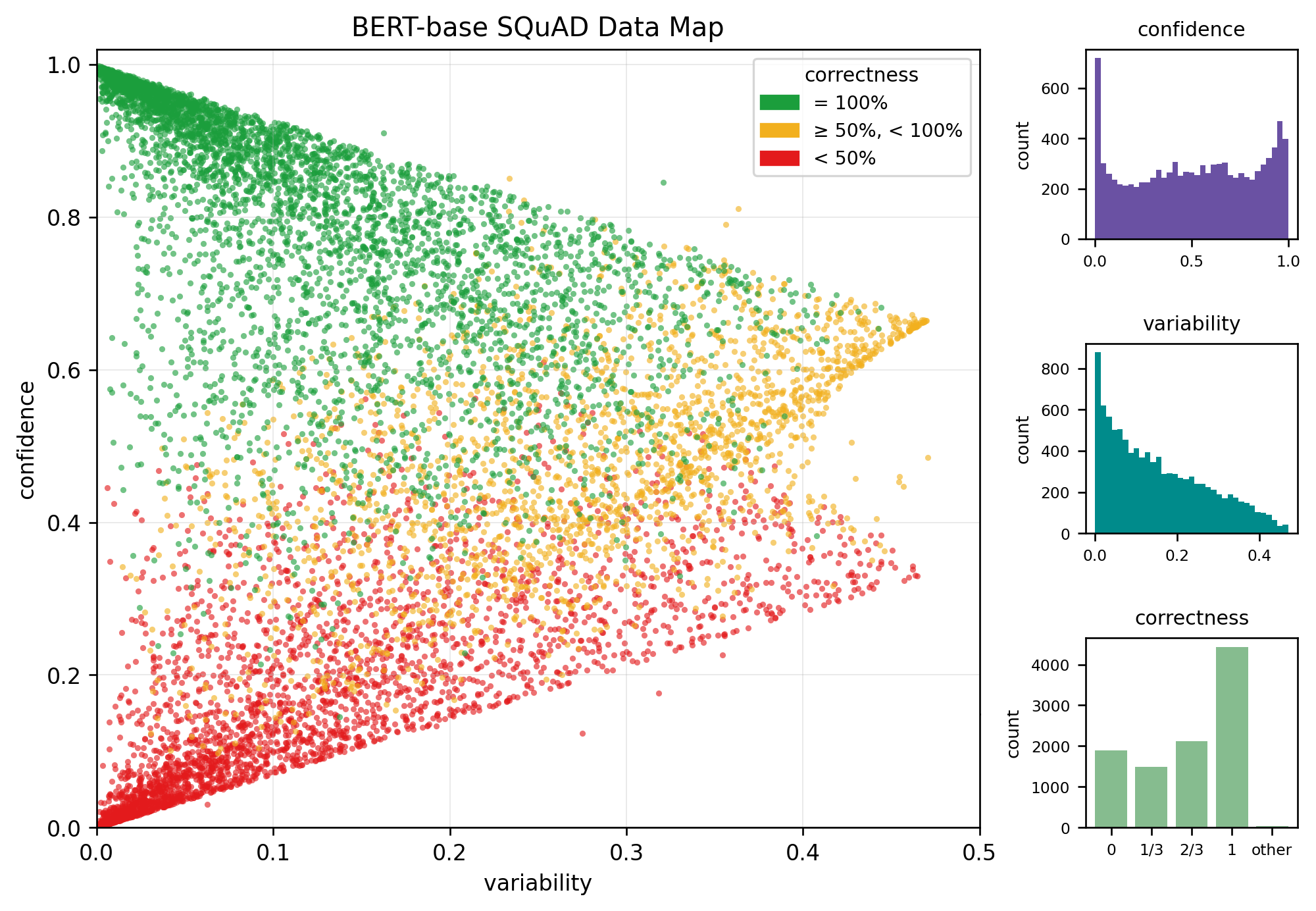}
\captionof{figure}{BERT-base SQuAD data map and dynamics marginals from the fixed 10K sample.}
\label{fig:profile-map-bert_base}
\vspace{1em}
\includegraphics[width=.81\linewidth]{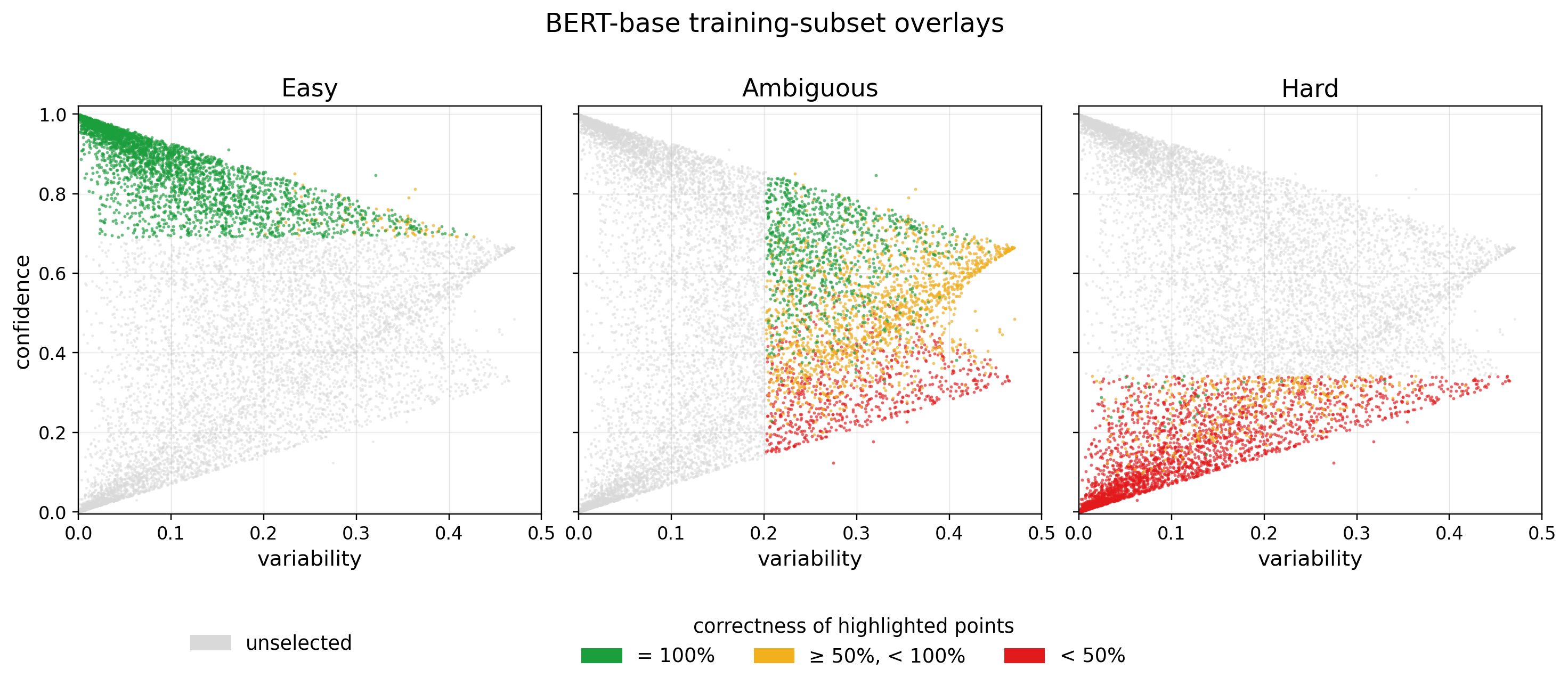}
\captionof{figure}{BERT-base global easy-, ambiguous-, and hard-ranked 33\% training subsets intersected with the same fixed 10K map sample. Gray points show the background sample; colored points are sampled members of the relevant global subset. Colors denote correctness categories and may be mixed within each panel. Highlight membership follows the globally ranked training subsets.}
\label{fig:profile-overlay-bert_base}

\par\endgroup

\clearpage
\noindent\begin{minipage}{\textwidth}

\centering
\small
\setlength{\tabcolsep}{5pt}
\begin{tabular}{lrrrr}
\toprule
Condition & AddSent EM & AddSent F1 & AddOneSent EM & AddOneSent F1 \\
\midrule
Full & \textbf{51.00 $\pm$ 0.23} & \textbf{57.66 $\pm$ 0.39} & \textbf{50.49 $\pm$ 0.60} & \textbf{57.10 $\pm$ 0.99} \\
Rand & 44.30 $\pm$ 1.28 & 51.27 $\pm$ 1.32 & 43.91 $\pm$ 1.34 & 50.47 $\pm$ 1.40 \\
Easy & 38.53 $\pm$ 1.00 & 43.70 $\pm$ 1.06 & 39.35 $\pm$ 1.51 & 44.19 $\pm$ 1.66 \\
Amb. & 47.42 $\pm$ 1.35 & 53.45 $\pm$ 1.41 & 47.73 $\pm$ 2.04 & 53.50 $\pm$ 2.14 \\
Hard & 38.37 $\pm$ 0.62 & 49.98 $\pm$ 1.04 & 35.96 $\pm$ 0.77 & 47.47 $\pm$ 1.15 \\
\bottomrule
\end{tabular}
\captionof{table}{Adversarial-only results for BERT-base. Best value in each column is bolded.}
\label{tab:model-specific-bert-base}

\end{minipage}

\vspace{12pt}
\begingroup

\centering
\includegraphics[width=.72\linewidth]{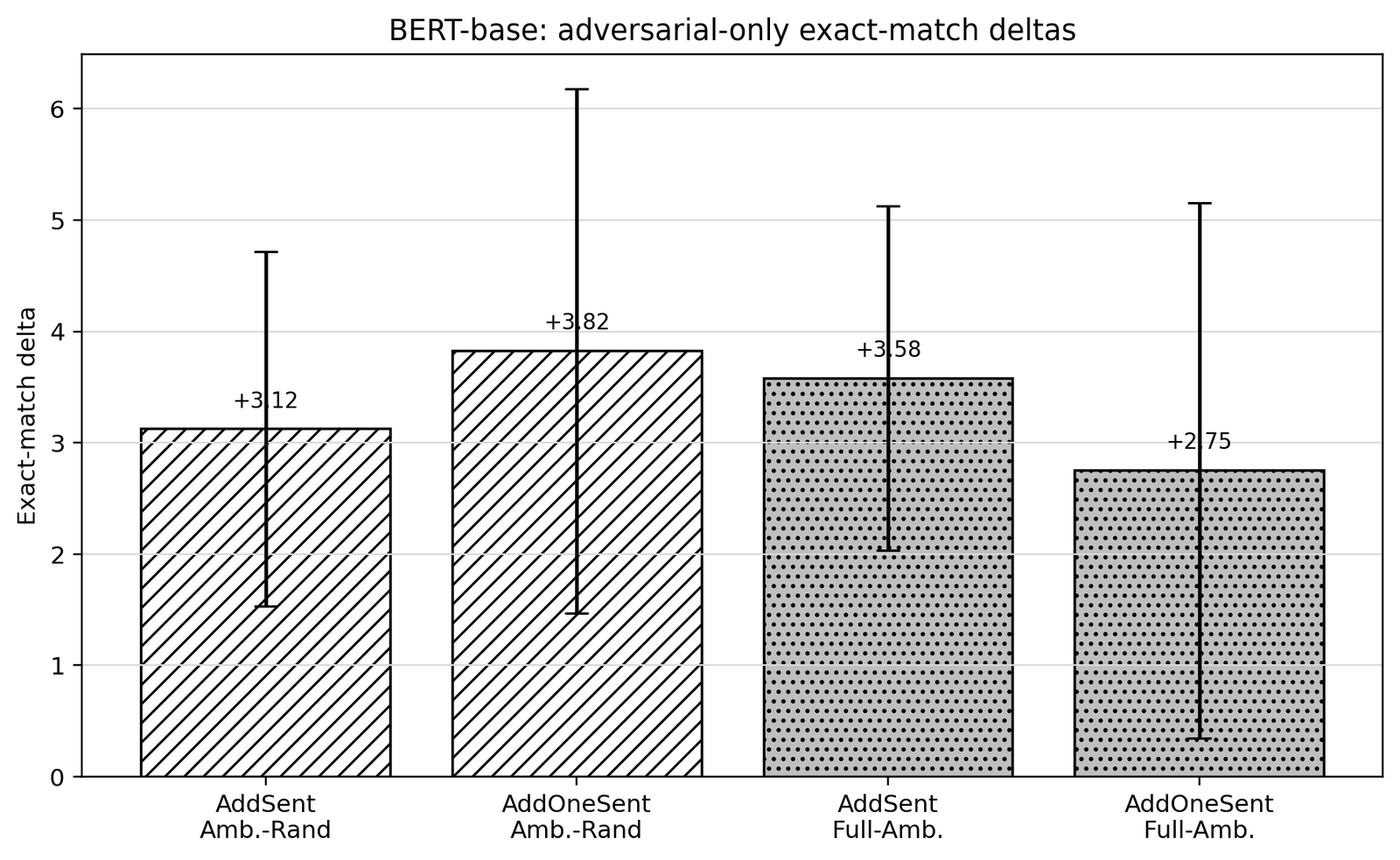}
\captionof{figure}{BERT-base adversarial-only exact-match delta bars. Amb.-Rand bars compare ambiguous-ranked 33\% training against the random 33\% baseline; Full-Amb. bars compare full-data training against ambiguous-ranked 33\%. Error bars are approximate 95\% confidence intervals from run-level standard errors.}
\label{fig:profile-delta-bert_base}

\par\endgroup

\clearpage
\begingroup

\raggedright\subsection{ELECTRA-base}
\centering
\includegraphics[width=.81\linewidth]{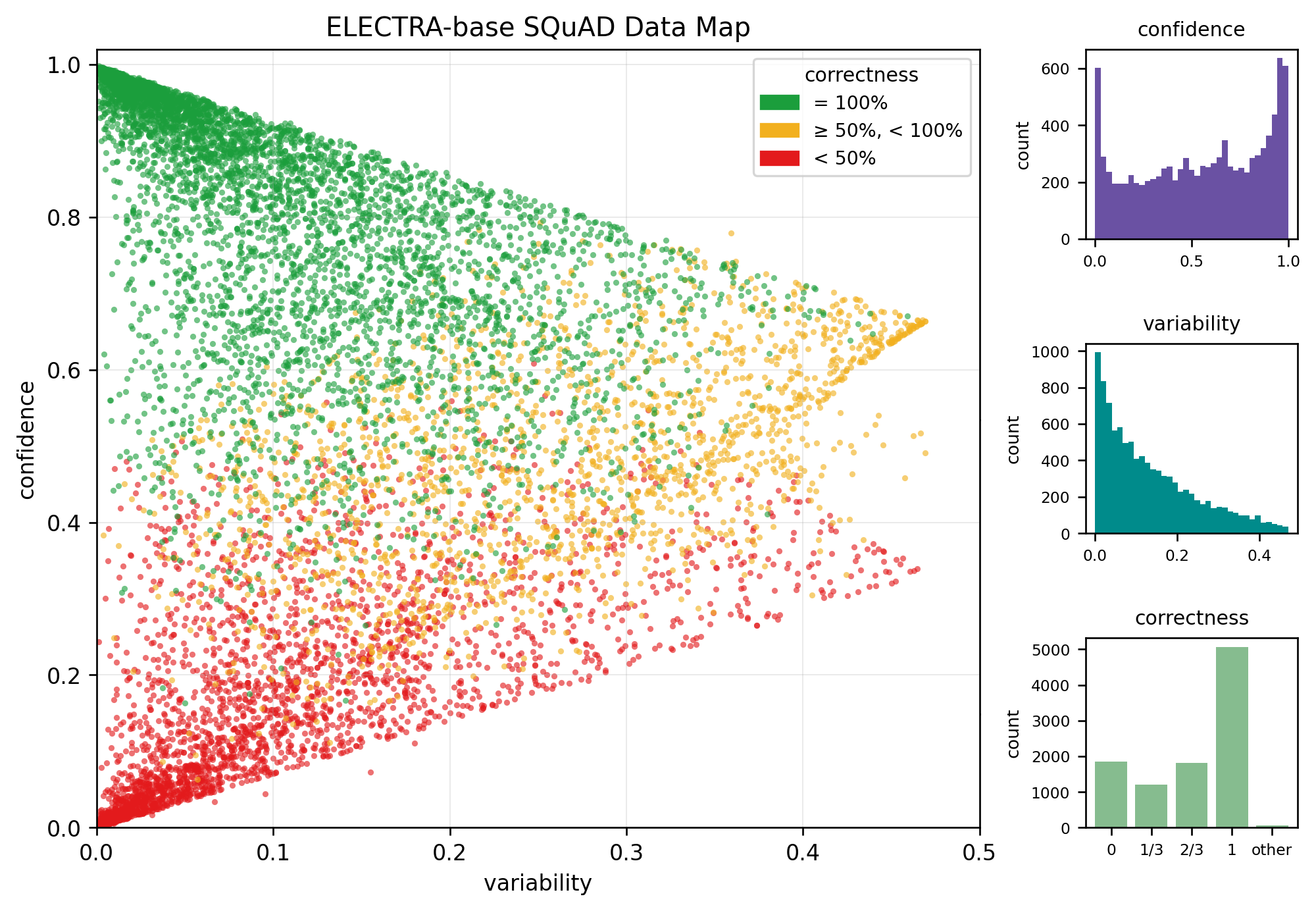}
\captionof{figure}{ELECTRA-base SQuAD data map and dynamics marginals from the fixed 10K sample.}
\label{fig:profile-map-electra_base}
\vspace{1em}
\includegraphics[width=.81\linewidth]{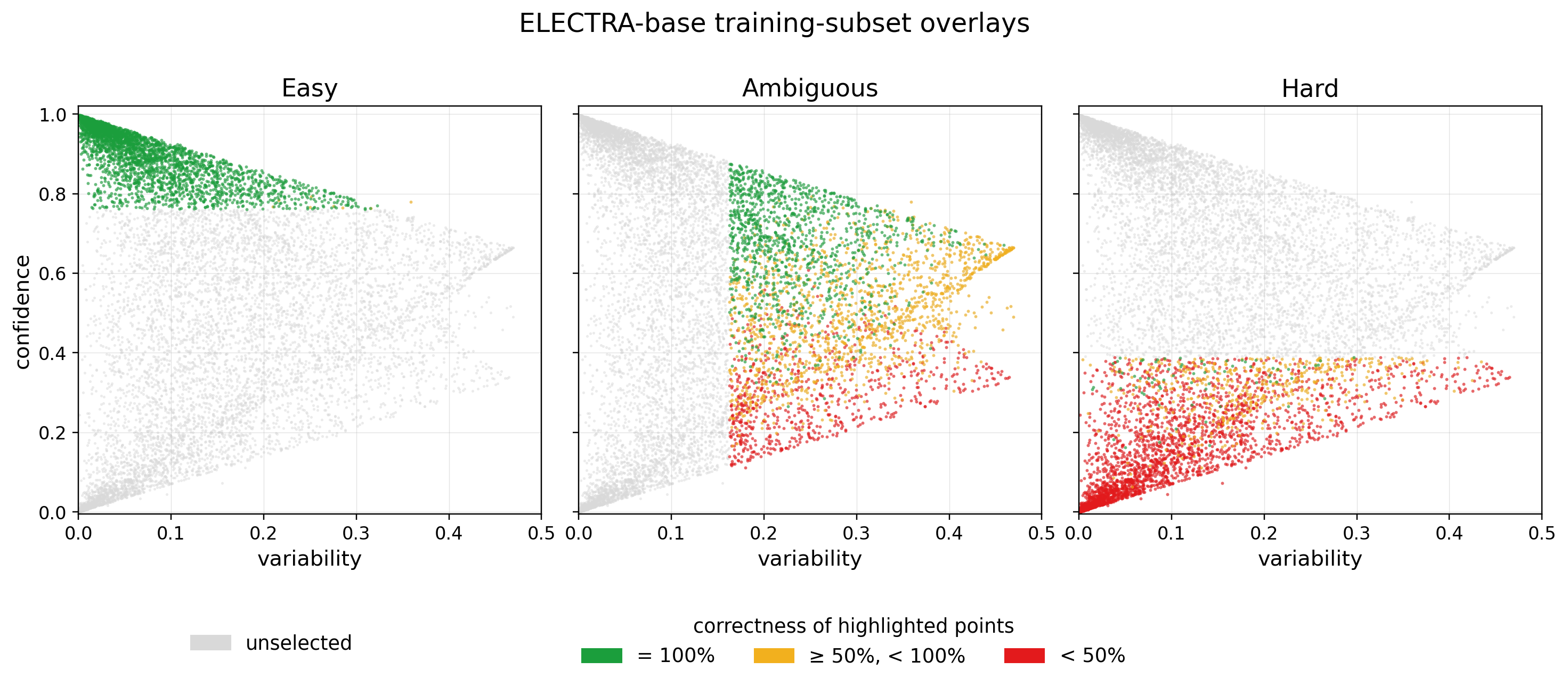}
\captionof{figure}{ELECTRA-base global easy-, ambiguous-, and hard-ranked 33\% training subsets intersected with the same fixed 10K map sample. Gray points show the background sample; colored points are sampled members of the relevant global subset. Colors denote correctness categories and may be mixed within each panel. Highlight membership follows the globally ranked training subsets.}
\label{fig:profile-overlay-electra_base}

\par\endgroup

\clearpage
\noindent\begin{minipage}{\textwidth}

\centering
\small
\setlength{\tabcolsep}{5pt}
\begin{tabular}{lrrrr}
\toprule
Condition & AddSent EM & AddSent F1 & AddOneSent EM & AddOneSent F1 \\
\midrule
Full & \textbf{62.57 $\pm$ 0.78} & \textbf{68.37 $\pm$ 1.21} & \textbf{62.35 $\pm$ 0.32} & \textbf{68.45 $\pm$ 0.71} \\
Rand & 54.85 $\pm$ 1.60 & 61.25 $\pm$ 1.70 & 54.36 $\pm$ 1.92 & 60.57 $\pm$ 2.18 \\
Easy & 49.45 $\pm$ 1.90 & 53.78 $\pm$ 1.80 & 50.28 $\pm$ 2.17 & 54.22 $\pm$ 2.19 \\
Amb. & 59.45 $\pm$ 1.39 & 65.27 $\pm$ 1.53 & 59.81 $\pm$ 1.20 & 65.47 $\pm$ 1.53 \\
Hard & 48.49 $\pm$ 0.09 & 60.15 $\pm$ 0.35 & 47.27 $\pm$ 0.55 & 59.07 $\pm$ 0.91 \\
\bottomrule
\end{tabular}
\captionof{table}{Adversarial-only results for ELECTRA-base. Best value in each column is bolded.}
\label{tab:model-specific-electra-base}

\end{minipage}

\vspace{12pt}
\begingroup

\centering
\includegraphics[width=.72\linewidth]{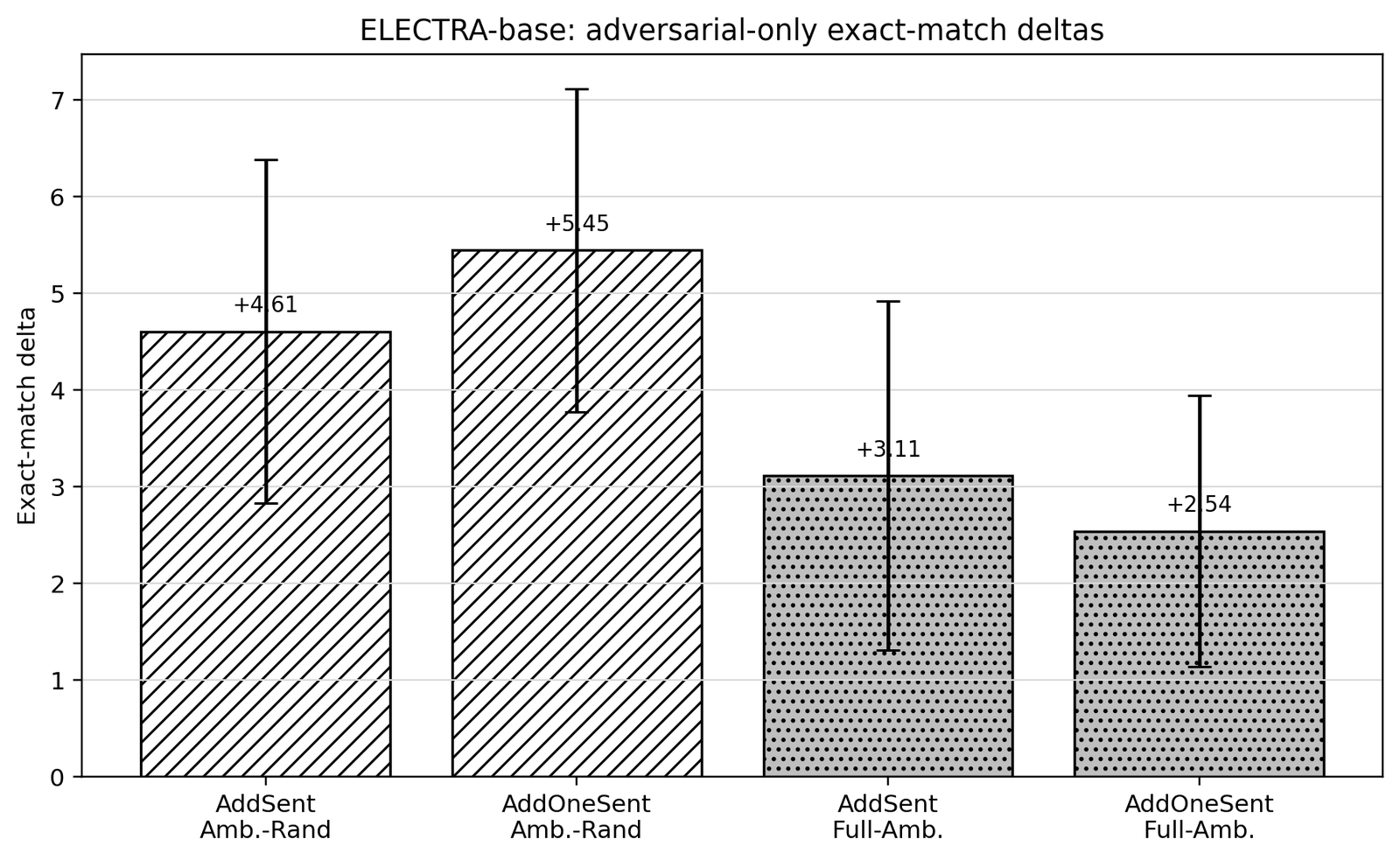}
\captionof{figure}{ELECTRA-base adversarial-only exact-match delta bars. Amb.-Rand bars compare ambiguous-ranked 33\% training against the random 33\% baseline; Full-Amb. bars compare full-data training against ambiguous-ranked 33\%. Error bars are approximate 95\% confidence intervals from run-level standard errors.}
\label{fig:profile-delta-electra_base}

\par\endgroup

\clearpage
\begingroup

\raggedright\subsection{RoBERTa-base}
\centering
\includegraphics[width=.81\linewidth]{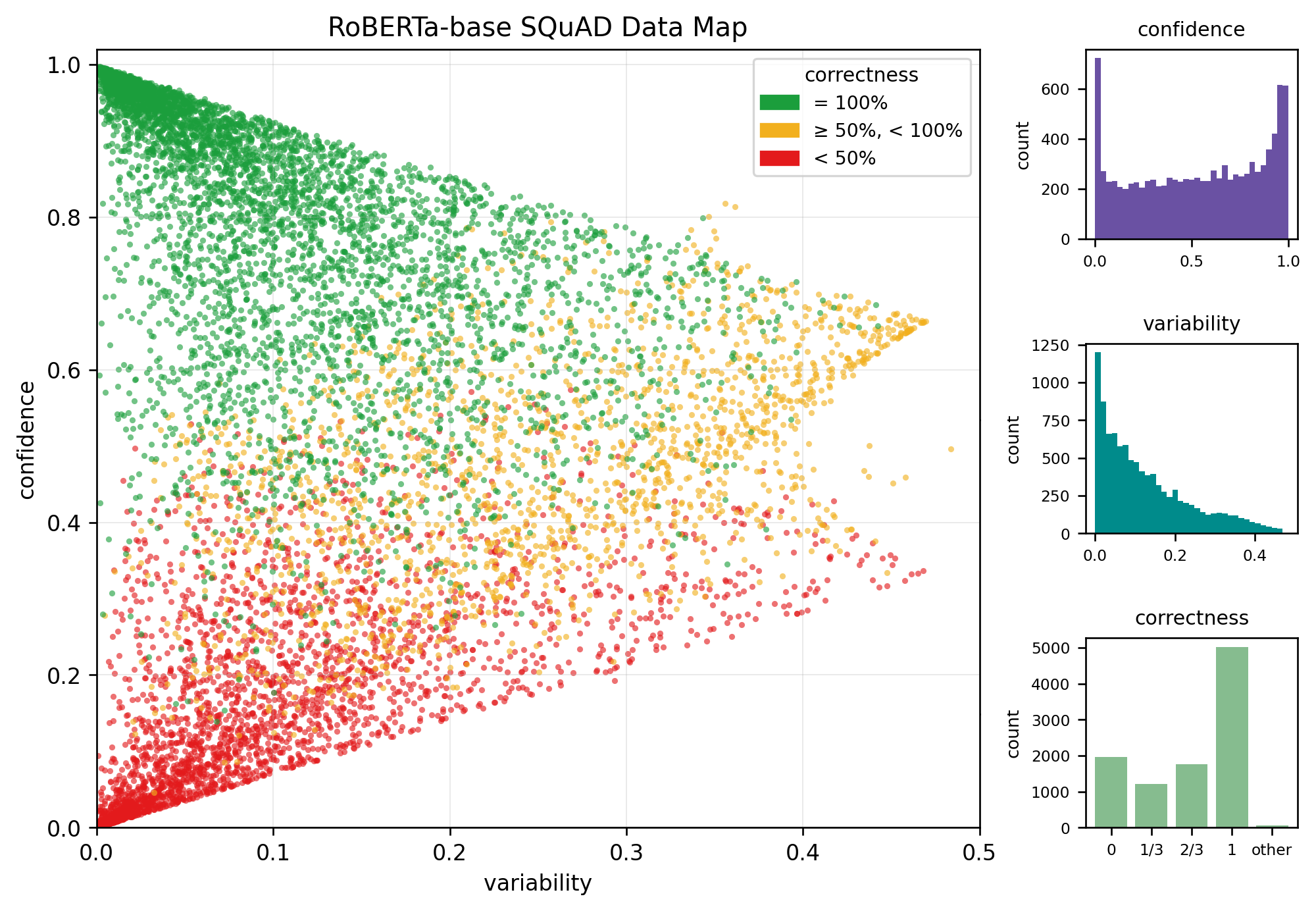}
\captionof{figure}{RoBERTa-base SQuAD data map and dynamics marginals from the fixed 10K sample.}
\label{fig:profile-map-roberta_base}
\vspace{1em}
\includegraphics[width=.81\linewidth]{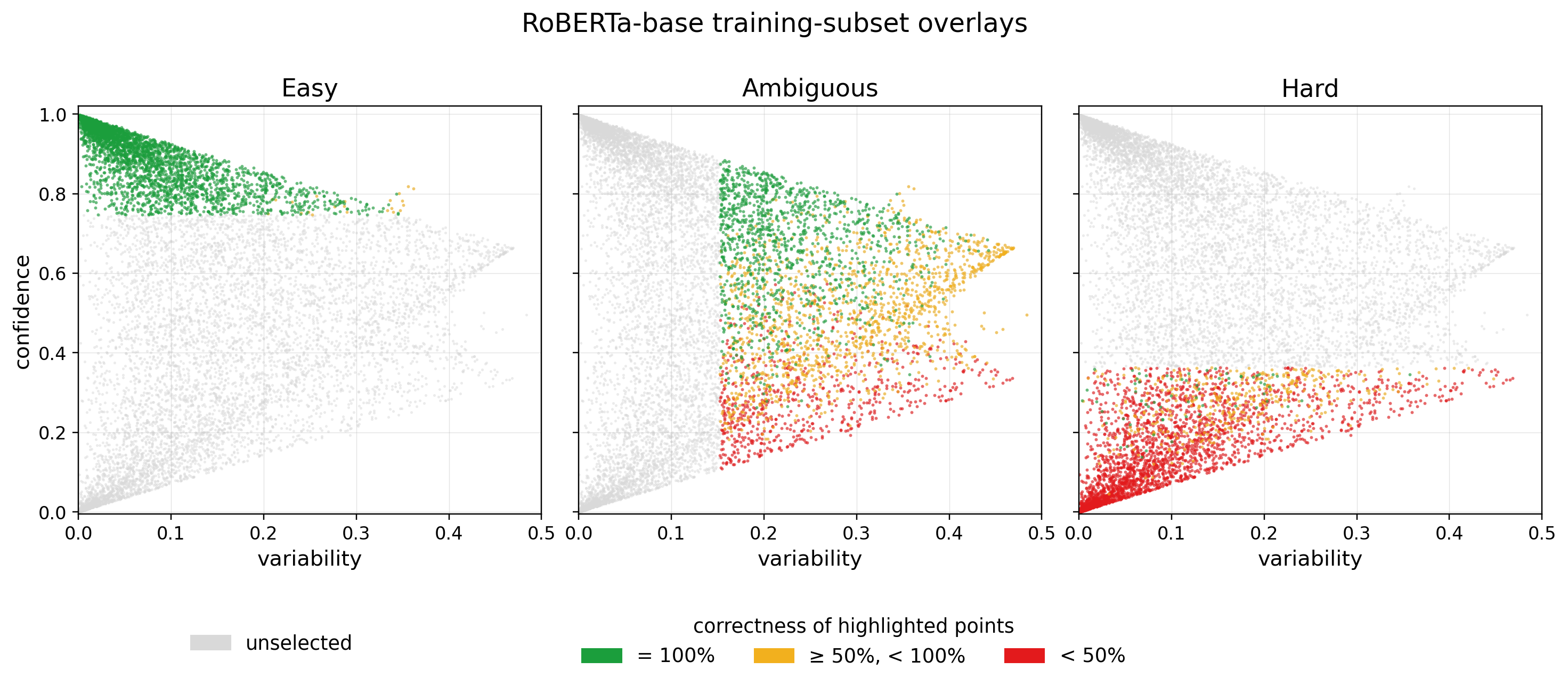}
\captionof{figure}{RoBERTa-base global easy-, ambiguous-, and hard-ranked 33\% training subsets intersected with the same fixed 10K map sample. Gray points show the background sample; colored points are sampled members of the relevant global subset. Colors denote correctness categories and may be mixed within each panel. Highlight membership follows the globally ranked training subsets.}
\label{fig:profile-overlay-roberta_base}

\par\endgroup

\clearpage
\noindent\begin{minipage}{\textwidth}

\centering
\small
\setlength{\tabcolsep}{5pt}
\begin{tabular}{lrrrr}
\toprule
Condition & AddSent EM & AddSent F1 & AddOneSent EM & AddOneSent F1 \\
\midrule
Full & \textbf{61.25 $\pm$ 0.52} & \textbf{67.67 $\pm$ 0.26} & \textbf{62.13 $\pm$ 0.13} & \textbf{68.23 $\pm$ 0.09} \\
Rand & 57.76 $\pm$ 1.45 & 64.14 $\pm$ 1.26 & 57.81 $\pm$ 1.31 & 63.95 $\pm$ 1.30 \\
Easy & 50.61 $\pm$ 0.99 & 54.95 $\pm$ 1.02 & 51.29 $\pm$ 0.60 & 55.42 $\pm$ 0.36 \\
Amb. & 59.90 $\pm$ 0.79 & 65.58 $\pm$ 1.02 & 59.76 $\pm$ 0.57 & 65.33 $\pm$ 0.74 \\
Hard & 49.27 $\pm$ 1.25 & 60.21 $\pm$ 1.57 & 48.84 $\pm$ 1.41 & 59.82 $\pm$ 1.53 \\
\bottomrule
\end{tabular}
\captionof{table}{Adversarial-only results for RoBERTa-base. Best value in each column is bolded.}
\label{tab:model-specific-roberta-base}

\end{minipage}

\vspace{12pt}
\begingroup

\centering
\includegraphics[width=.72\linewidth]{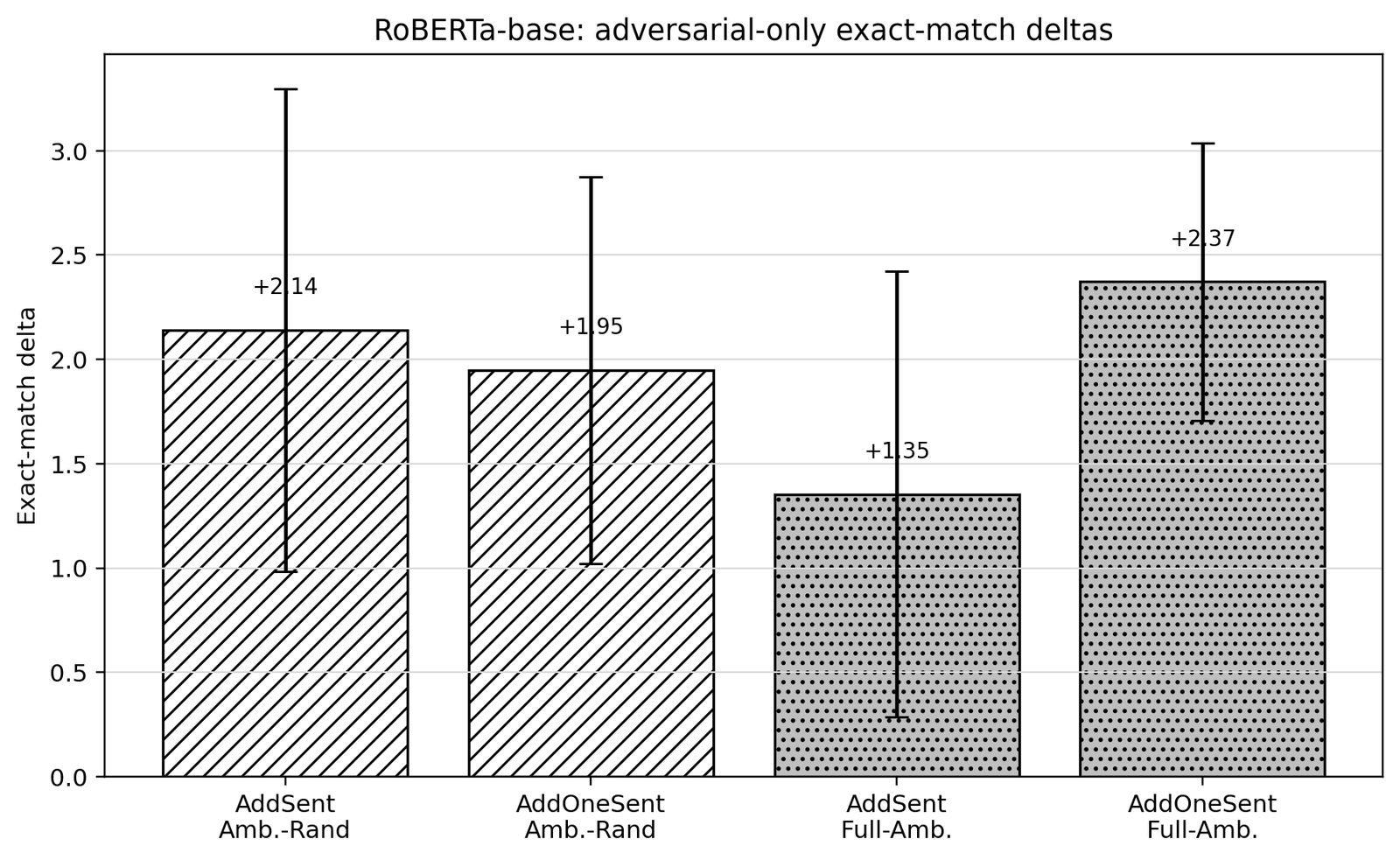}
\captionof{figure}{RoBERTa-base adversarial-only exact-match delta bars. Amb.-Rand bars compare ambiguous-ranked 33\% training against the random 33\% baseline; Full-Amb. bars compare full-data training against ambiguous-ranked 33\%. Error bars are approximate 95\% confidence intervals from run-level standard errors.}
\label{fig:profile-delta-roberta_base}

\par\endgroup

\clearpage
\begingroup

\raggedright\subsection{DistilBERT}
\centering
\includegraphics[width=.81\linewidth]{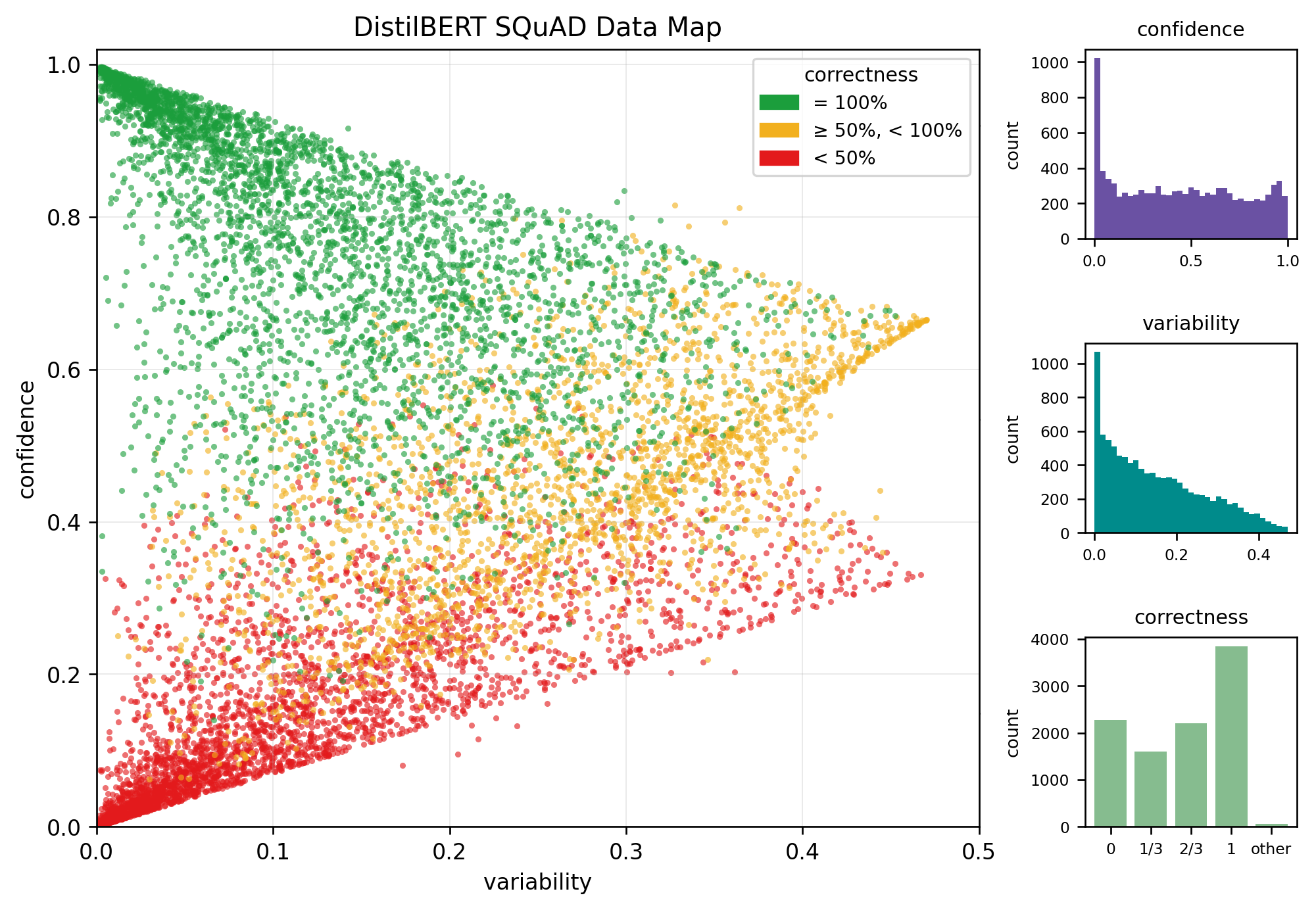}
\captionof{figure}{DistilBERT SQuAD data map and dynamics marginals from the fixed 10K sample.}
\label{fig:profile-map-distilbert}
\vspace{1em}
\includegraphics[width=.81\linewidth]{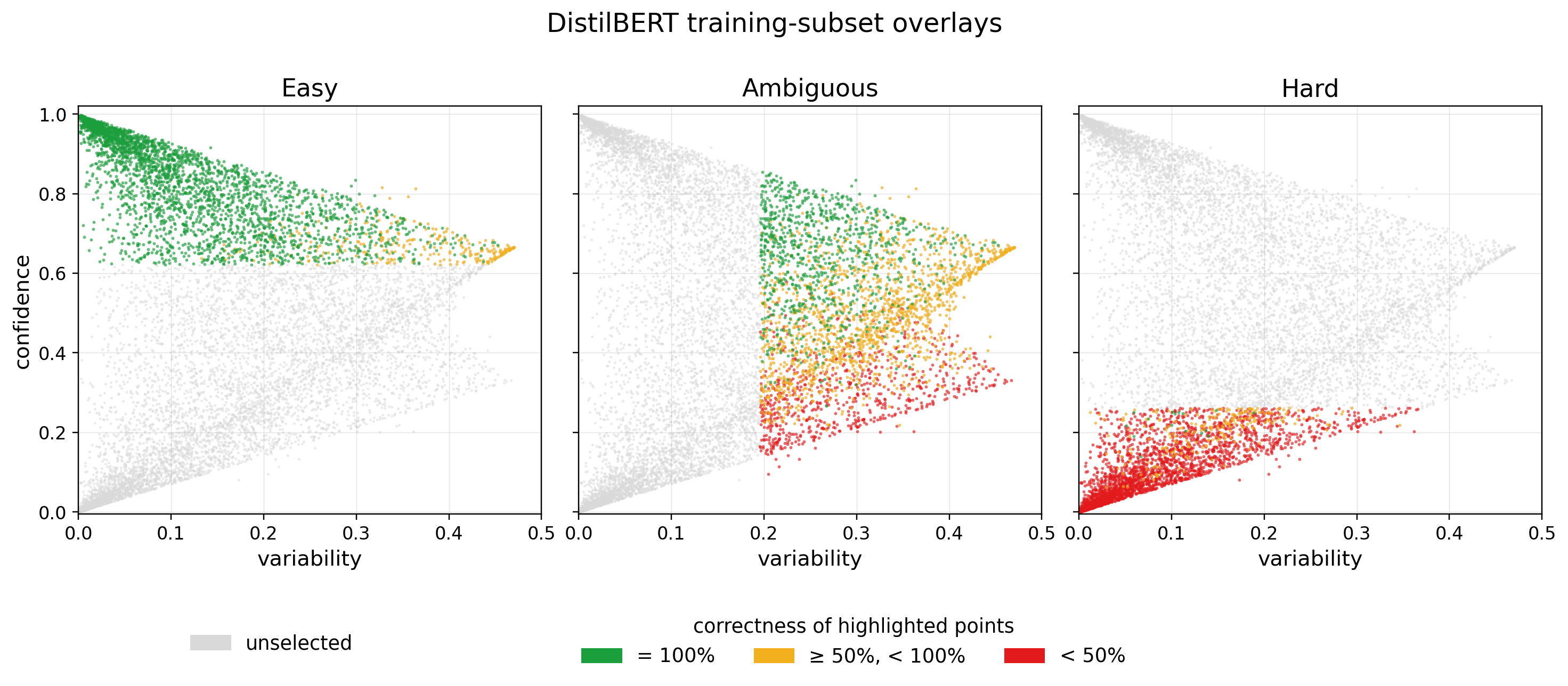}
\captionof{figure}{DistilBERT global easy-, ambiguous-, and hard-ranked 33\% training subsets intersected with the same fixed 10K map sample. Gray points show the background sample; colored points are sampled members of the relevant global subset. Colors denote correctness categories and may be mixed within each panel. Highlight membership follows the globally ranked training subsets.}
\label{fig:profile-overlay-distilbert}

\par\endgroup

\clearpage
\noindent\begin{minipage}{\textwidth}

\centering
\small
\setlength{\tabcolsep}{5pt}
\begin{tabular}{lrrrr}
\toprule
Condition & AddSent EM & AddSent F1 & AddOneSent EM & AddOneSent F1 \\
\midrule
Full & \textbf{46.08 $\pm$ 1.14} & \textbf{53.15 $\pm$ 1.17} & \textbf{45.70 $\pm$ 1.48} & \textbf{52.50 $\pm$ 1.54} \\
Rand & 38.34 $\pm$ 1.41 & 45.61 $\pm$ 1.33 & 37.78 $\pm$ 1.27 & 44.80 $\pm$ 1.23 \\
Easy & 32.30 $\pm$ 0.57 & 37.64 $\pm$ 0.90 & 32.70 $\pm$ 0.73 & 38.24 $\pm$ 1.20 \\
Amb. & 40.08 $\pm$ 1.22 & 46.31 $\pm$ 1.27 & 40.24 $\pm$ 1.27 & 46.37 $\pm$ 1.58 \\
Hard & 35.26 $\pm$ 1.26 & 47.02 $\pm$ 1.39 & 33.55 $\pm$ 1.44 & 45.29 $\pm$ 1.60 \\
\bottomrule
\end{tabular}
\captionof{table}{Adversarial-only results for DistilBERT. Best value in each column is bolded.}
\label{tab:model-specific-distilbert}

\end{minipage}

\vspace{12pt}
\begingroup

\centering
\includegraphics[width=.72\linewidth]{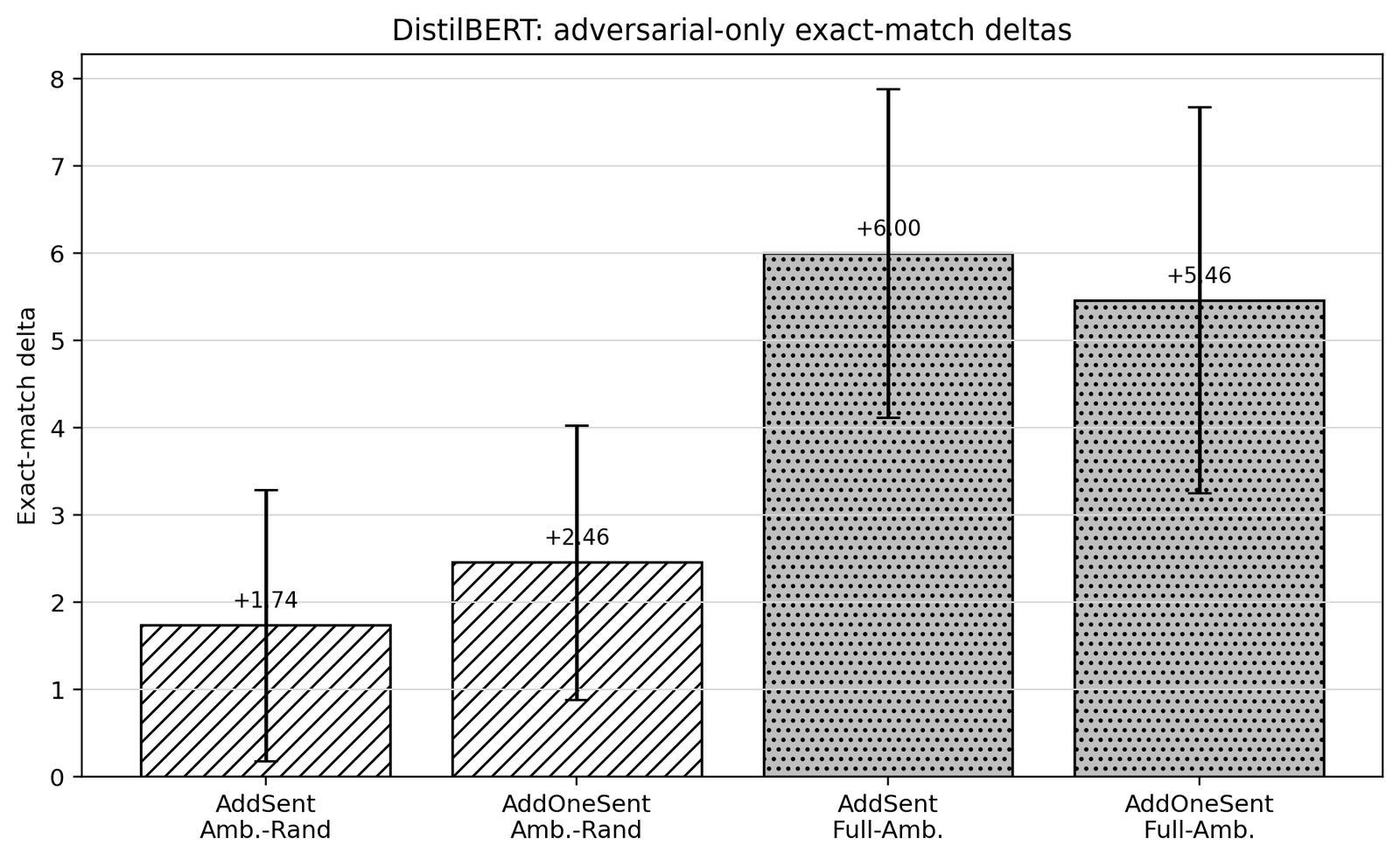}
\captionof{figure}{DistilBERT adversarial-only exact-match delta bars. Amb.-Rand bars compare ambiguous-ranked 33\% training against the random 33\% baseline; Full-Amb. bars compare full-data training against ambiguous-ranked 33\%. Error bars are approximate 95\% confidence intervals from run-level standard errors.}
\label{fig:profile-delta-distilbert}

\par\endgroup

\clearpage
\begingroup

\raggedright\subsection{ALBERT-base}
\centering
\includegraphics[width=.81\linewidth]{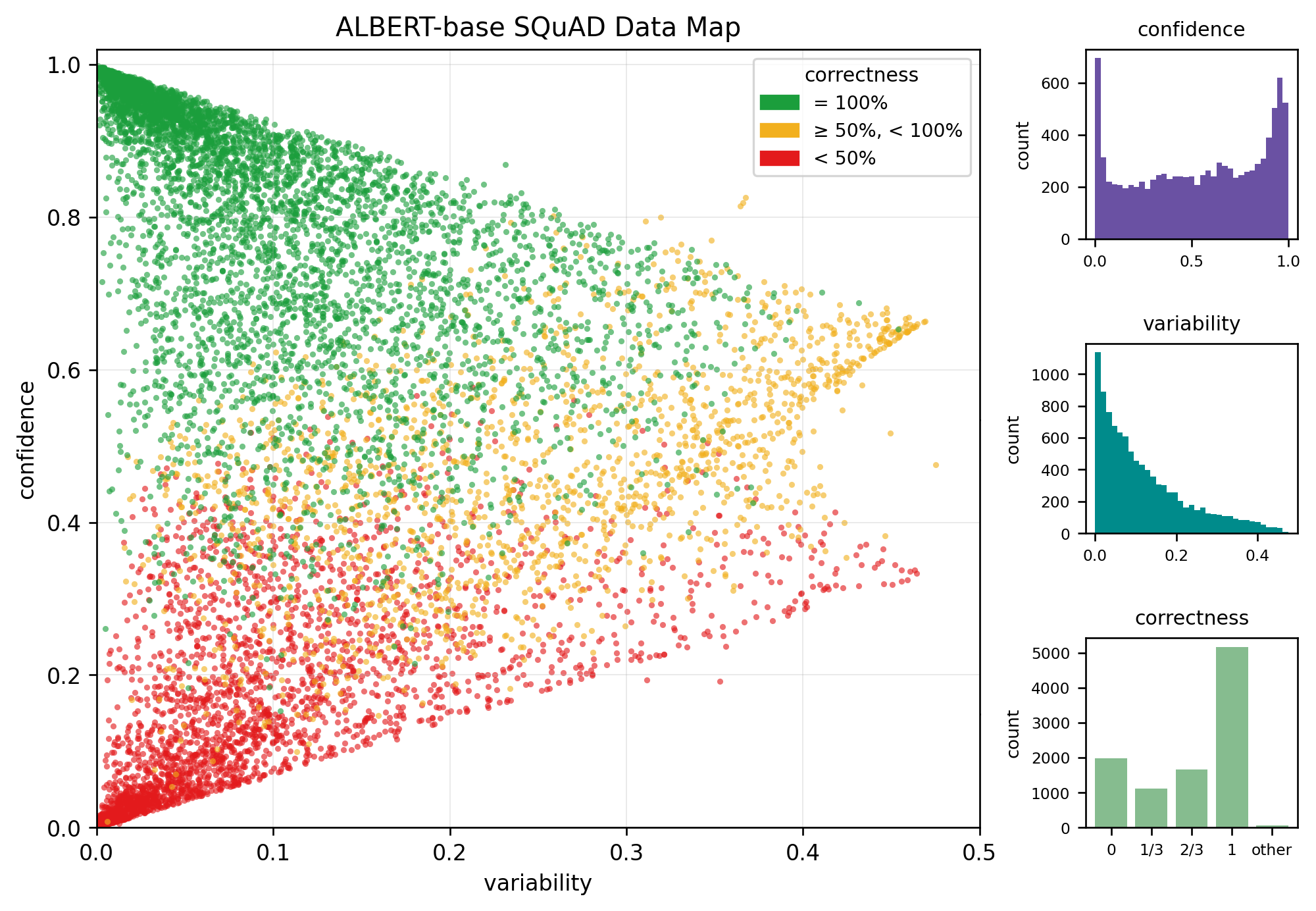}
\captionof{figure}{ALBERT-base SQuAD data map and dynamics marginals from the fixed 10K sample.}
\label{fig:profile-map-albert_base}
\vspace{1em}
\includegraphics[width=.81\linewidth]{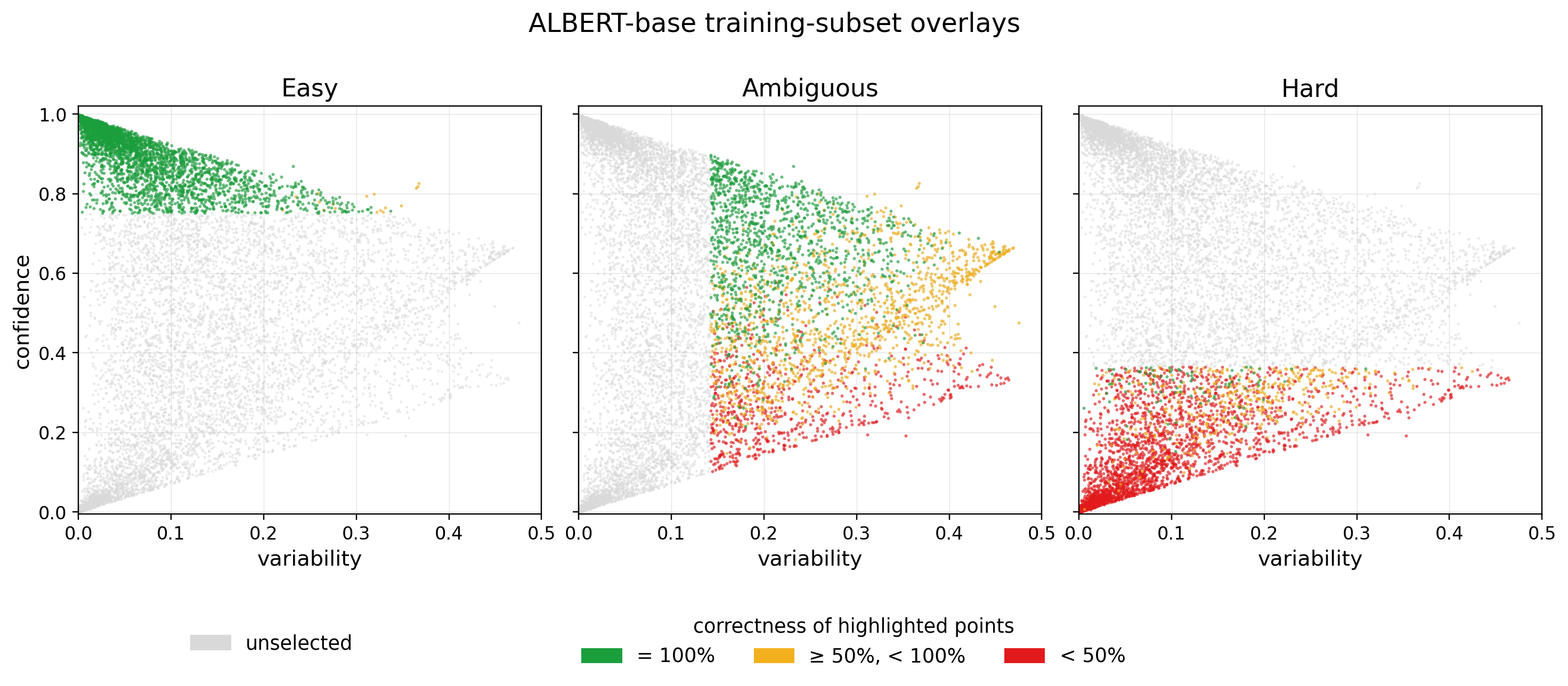}
\captionof{figure}{ALBERT-base global easy-, ambiguous-, and hard-ranked 33\% training subsets intersected with the same fixed 10K map sample. Gray points show the background sample; colored points are sampled members of the relevant global subset. Colors denote correctness categories and may be mixed within each panel. Highlight membership follows the globally ranked training subsets.}
\label{fig:profile-overlay-albert_base}

\par\endgroup

\clearpage
\noindent\begin{minipage}{\textwidth}

\centering
\small
\setlength{\tabcolsep}{5pt}
\begin{tabular}{lrrrr}
\toprule
Condition & AddSent EM & AddSent F1 & AddOneSent EM & AddOneSent F1 \\
\midrule
Full & \textbf{55.40 $\pm$ 1.04} & \textbf{61.69 $\pm$ 0.78} & \textbf{55.65 $\pm$ 0.92} & \textbf{62.00 $\pm$ 0.90} \\
Rand & 49.77 $\pm$ 1.50 & 56.03 $\pm$ 1.50 & 49.99 $\pm$ 1.45 & 55.99 $\pm$ 1.48 \\
Easy & 41.84 $\pm$ 1.40 & 46.56 $\pm$ 1.48 & 41.00 $\pm$ 1.73 & 45.62 $\pm$ 1.81 \\
Amb. & 52.90 $\pm$ 2.06 & 58.05 $\pm$ 1.98 & 53.66 $\pm$ 1.73 & 58.25 $\pm$ 1.77 \\
Hard & 43.53 $\pm$ 1.73 & 53.94 $\pm$ 1.45 & 43.03 $\pm$ 1.80 & 53.53 $\pm$ 1.79 \\
\bottomrule
\end{tabular}
\captionof{table}{Adversarial-only results for ALBERT-base. Best value in each column is bolded.}
\label{tab:model-specific-albert-base}

\end{minipage}

\vspace{12pt}
\begingroup

\centering
\includegraphics[width=.72\linewidth]{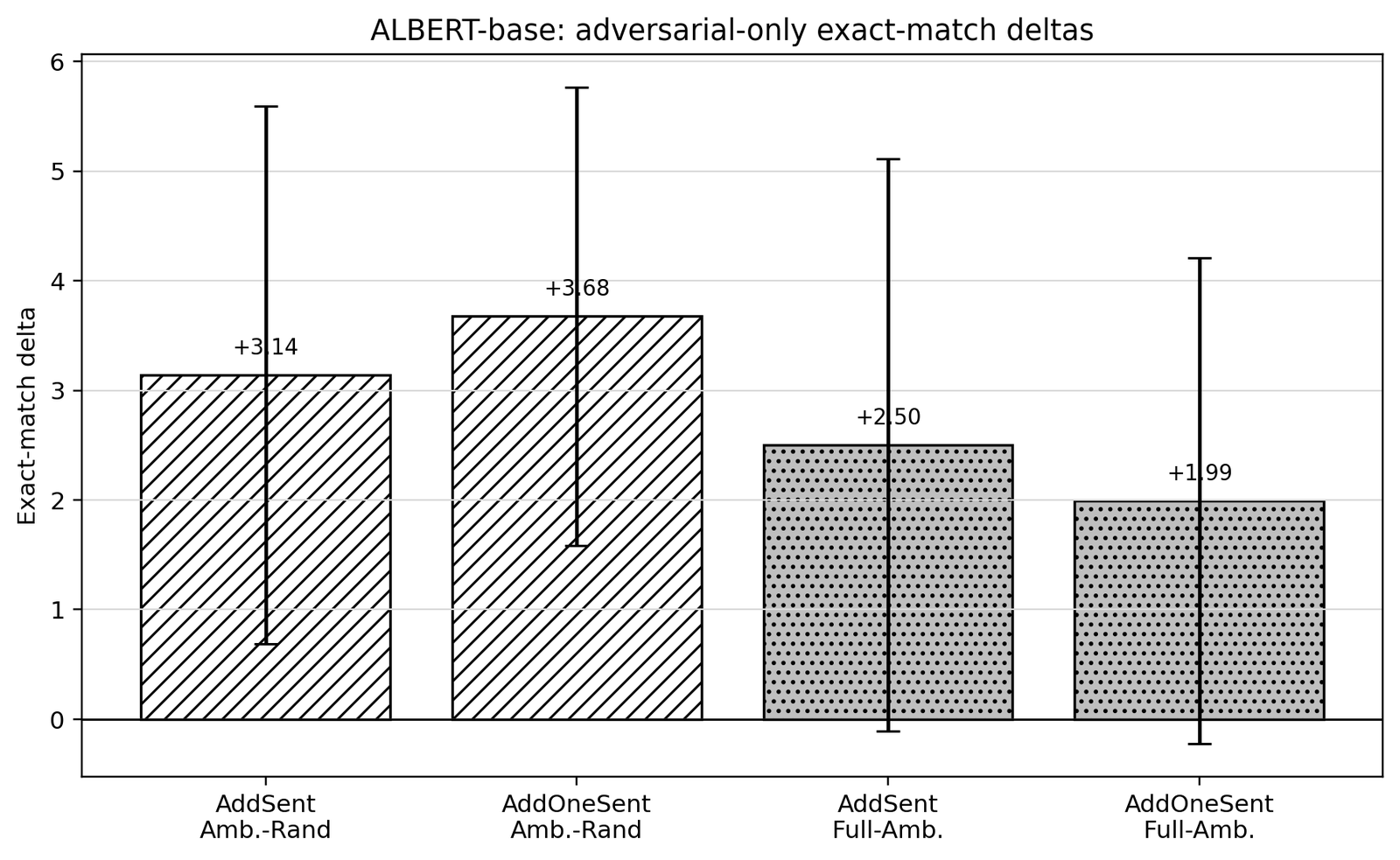}
\captionof{figure}{ALBERT-base adversarial-only exact-match delta bars. Amb.-Rand bars compare ambiguous-ranked 33\% training against the random 33\% baseline; Full-Amb. bars compare full-data training against ambiguous-ranked 33\%. Error bars are approximate 95\% confidence intervals from run-level standard errors.}
\label{fig:profile-delta-albert_base}

\par\endgroup

\clearpage
\begingroup

\raggedright\subsection{DeBERTa-v3-base}
\centering
\includegraphics[width=.81\linewidth]{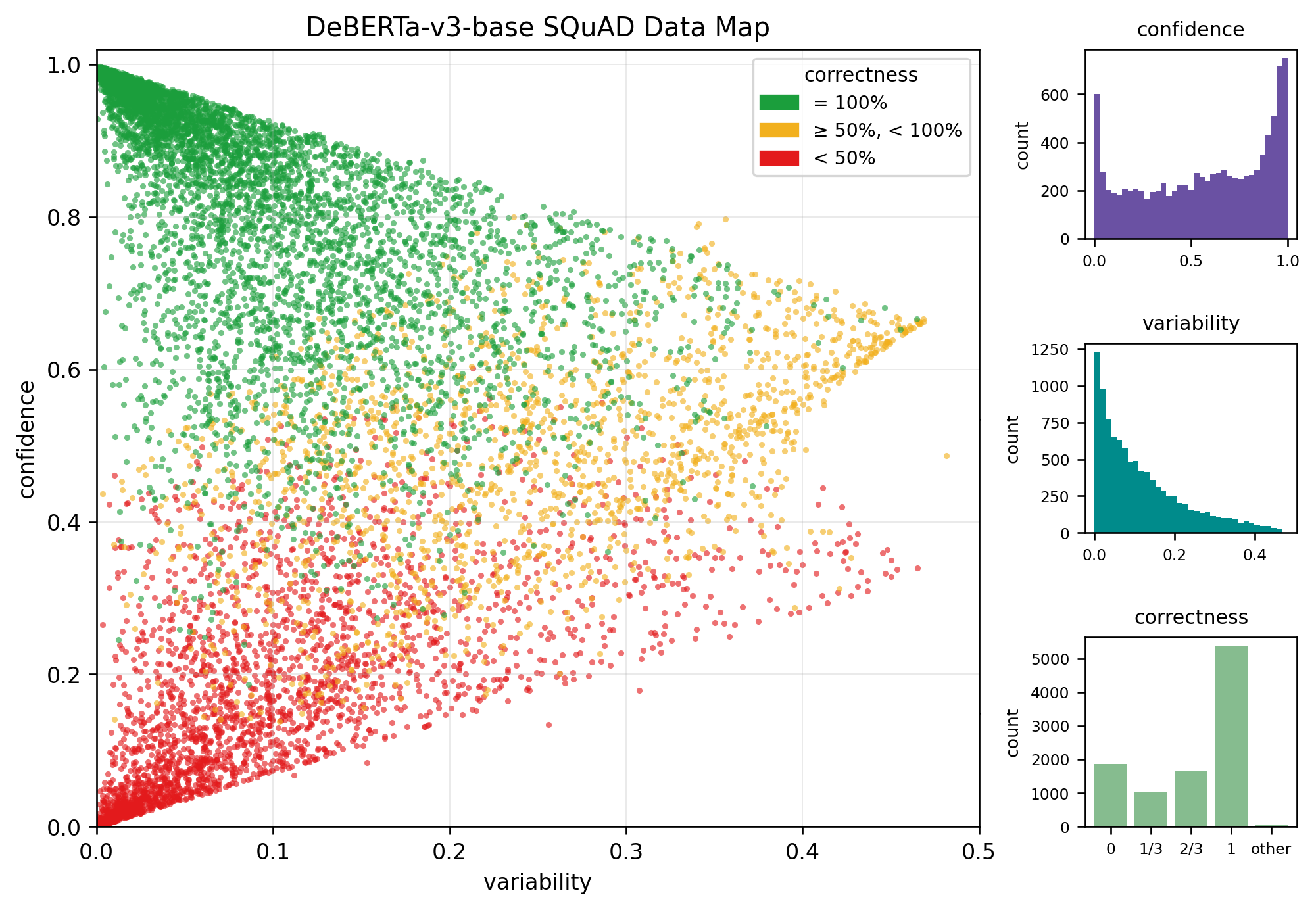}
\captionof{figure}{DeBERTa-v3-base SQuAD data map and dynamics marginals from the fixed 10K sample.}
\label{fig:profile-map-deberta_v3_base}
\vspace{1em}
\includegraphics[width=.81\linewidth]{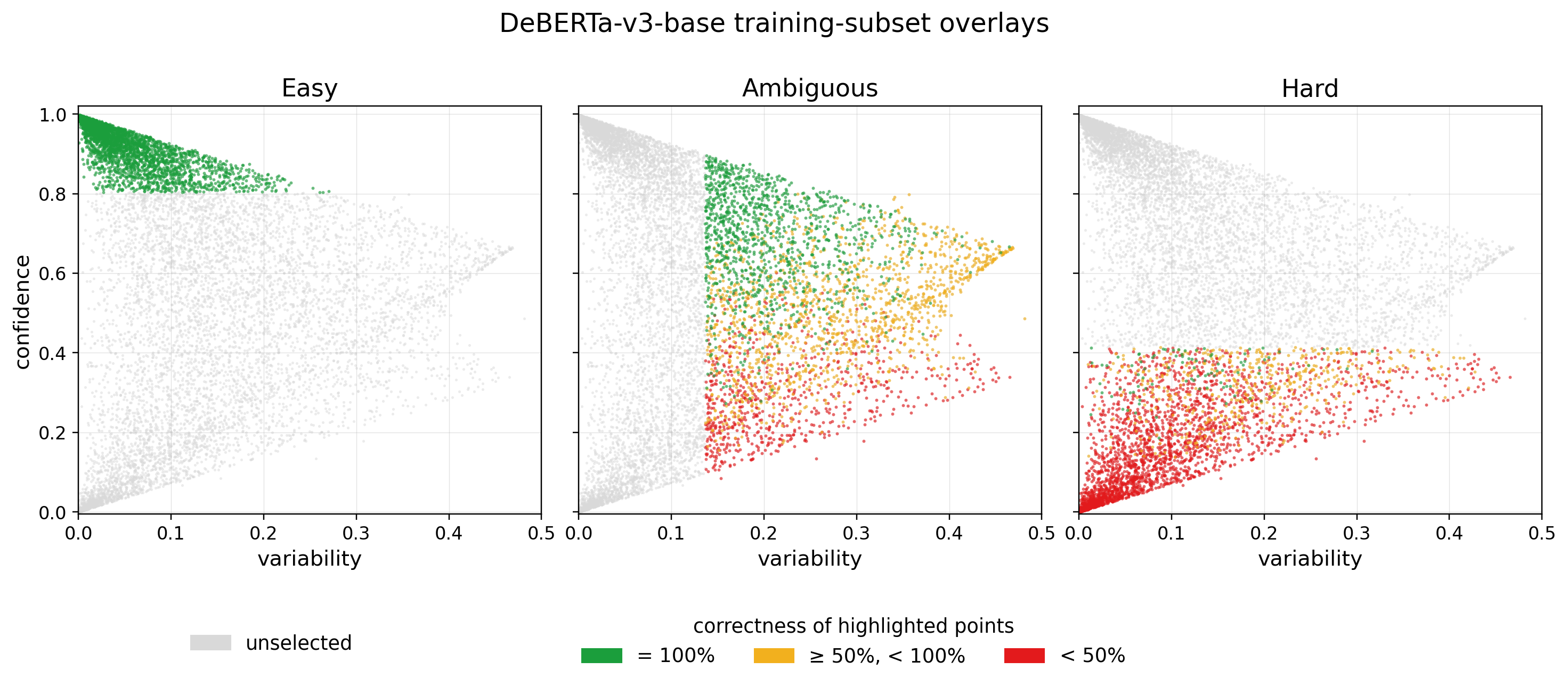}
\captionof{figure}{DeBERTa-v3-base global easy-, ambiguous-, and hard-ranked 33\% training subsets intersected with the same fixed 10K map sample. Gray points show the background sample; colored points are sampled members of the relevant global subset. Colors denote correctness categories and may be mixed within each panel. Highlight membership follows the globally ranked training subsets.}
\label{fig:profile-overlay-deberta_v3_base}

\par\endgroup

\clearpage
\noindent\begin{minipage}{\textwidth}

\centering
\small
\setlength{\tabcolsep}{5pt}
\begin{tabular}{lrrrr}
\toprule
Condition & AddSent EM & AddSent F1 & AddOneSent EM & AddOneSent F1 \\
\midrule
Full & 74.67 $\pm$ 0.90 & 80.03 $\pm$ 1.14 & 75.01 $\pm$ 0.73 & 80.11 $\pm$ 0.80 \\
Rand & 71.84 $\pm$ 0.94 & 78.15 $\pm$ 0.83 & 71.83 $\pm$ 1.05 & 77.84 $\pm$ 1.09 \\
Easy & 60.51 $\pm$ 2.19 & 64.85 $\pm$ 2.24 & 61.33 $\pm$ 2.27 & 65.60 $\pm$ 2.36 \\
Amb. & \textbf{74.99 $\pm$ 1.11} & \textbf{80.33 $\pm$ 1.29} & \textbf{75.35 $\pm$ 1.44} & \textbf{80.35 $\pm$ 1.65} \\
Hard & 59.99 $\pm$ 0.26 & 73.35 $\pm$ 0.29 & 58.20 $\pm$ 0.77 & 72.11 $\pm$ 1.04 \\
\bottomrule
\end{tabular}
\captionof{table}{Adversarial-only results for DeBERTa-v3-base. Best value in each column is bolded.}
\label{tab:model-specific-deberta-v3-base}

\end{minipage}

\vspace{12pt}
\begingroup

\centering
\includegraphics[width=.72\linewidth]{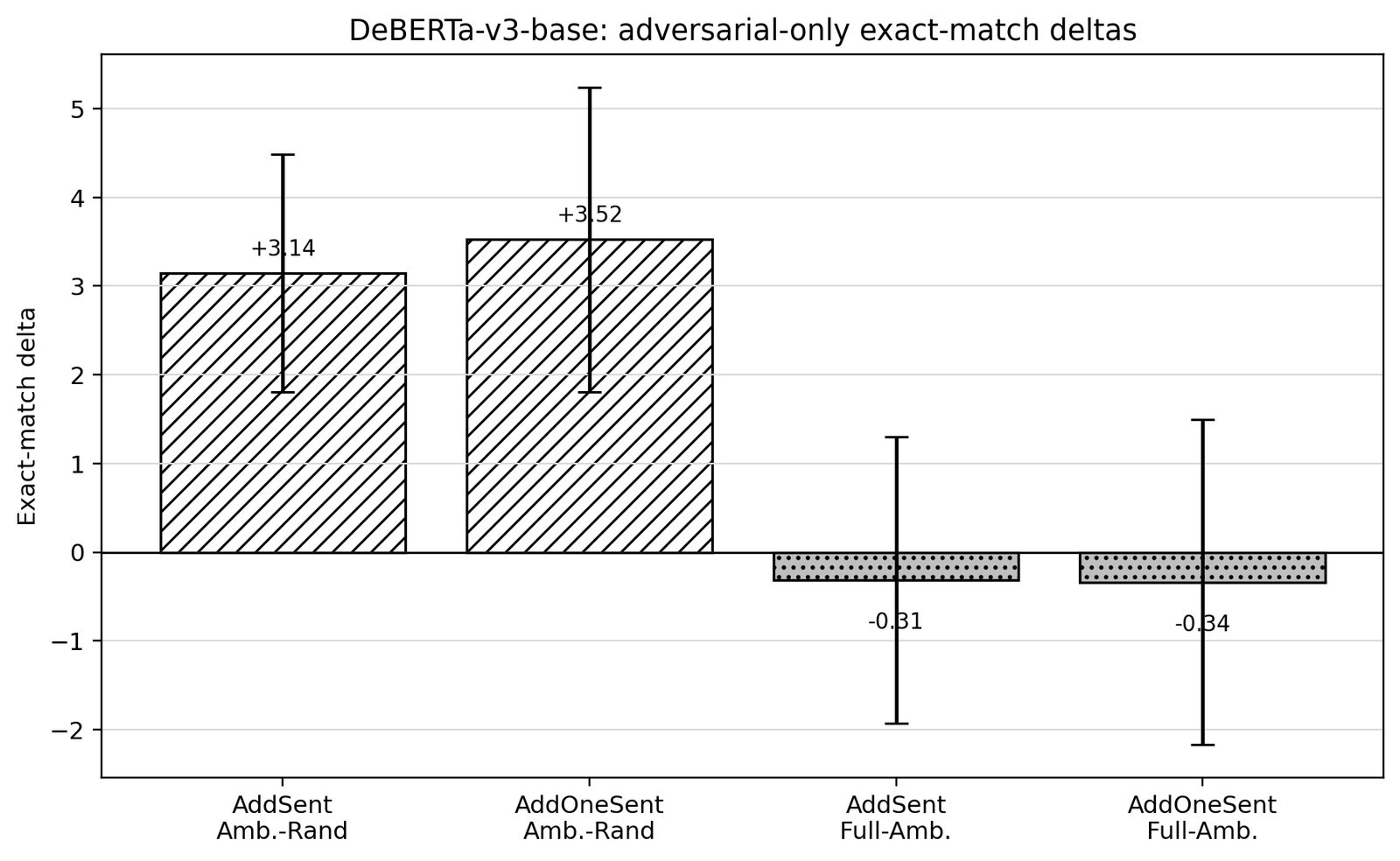}
\captionof{figure}{DeBERTa-v3-base adversarial-only exact-match delta bars. Amb.-Rand bars compare ambiguous-ranked 33\% training against the random 33\% baseline; Full-Amb. bars compare full-data training against ambiguous-ranked 33\%. Error bars are approximate 95\% confidence intervals from run-level standard errors.}
\label{fig:profile-delta-deberta_v3_base}

\par\endgroup

\end{document}